\newif\iffinal
\finaltrue 
\documentclass[12pt, a4paper]{article}
\usepackage{graphicx}
\usepackage{color}
\usepackage{hyperref}
\usepackage{xspace}
\newcommand{\delftacopter}{\emph{DelftaCopter}\xspace}
\newcommand{\slamdunk}{S.L.A.M.dunk\xspace}
\newcommand{\includematlab}[1]{\includegraphics[scale=0.8]{#1}}

\date{September 28, 2016}

\begin{document}

\title{Design, Control and Visual Navigation of the \delftacopter}

\author{ C. De Wagter\footnote{Email address: c.dewagter@tudelft.nl, Micro Air Vehicle Lab, Kluyverweg 1, 2629HS Delft, the Netherlands.}, R. Ruijsink\footnote{Email address: info@ruijsink.nl}, E.J.J. Smeur\footnote{Email address: e.j.j.smeur@tudelft.nl}, K. van Hecke, \\F. van Tienen, E. v.d. Horst and B. Remes\footnote{Email address: b.d.w.remes@tudelft.nl}}

\maketitle

\begin{abstract}
To participate in the \emph{Outback Medical Express UAV Challenge 2016}, a vehicle was designed and tested that can hover precisely, take-off and land vertically, fly fast forward efficiently and use computer vision to locate a person and a suitable landing location.
A rotor blade was designed that can deliver sufficient thrust in hover, while still being efficient in fast forward flight. Energy measurements and windtunnel tests were performed.
A rotor-head and corresponding control algorithms were developed to allow transitioning flight with the non-conventional rotor dynamics. Dedicated electronics were designed that meet vehicle needs and regulations to allow safe flight beyond visual line of sight.
Vision based search and guidance algorithms were developed and tested. Flight tests and a competition participation illustrate the applicability of the \delftacopter concept.
\end{abstract}


\section*{Nomenclature}
\noindent\begin{tabular}{@{}lcl@{}}
\textit{CASA}  &=& Civil Aviation Safety Authority \\
\textit{IMU}  &=& Inertial Measurement Unit \\
\textit{GPS}  &=& Global Positioning System \\
\textit{UAV}  &=& Unmanned Air Vehicle \\
\textit{VTOL}  &=& Vetical Take-Off and Landing \\
\end{tabular} \\

\section{Introduction} \label{section:Introduction}

It has always been a goal in aviation to design aircraft which are efficient and controllable from hover to very fast forward flight. Unfortunately requirements for fast and slow flight are very contradictory \cite{anderson1999aircraft}. While hybrid aircraft have existed for a long time, the search for the ultimate combination is still ongoing \cite{wong2007attitude,imav2016:y_ke_et_al,imav2011:m_itasse_et_al,imav2014:m_schutt_et_al,PHUNG2013212,ccetinsoy2012design,escareno2007modeling}.
With the advent of unmanned air vehicles, several hybrid aircraft concepts that were previously impracticable have gained new interest \cite{de2013multi,escareno2007modeling}.

The applications of aircraft with combined efficient long-range flight and hovering capabilities are numerous. Typical examples are operation from ships or within forests. 

\begin{figure}[hbt]
\centering
\includegraphics[width=10.4cm]{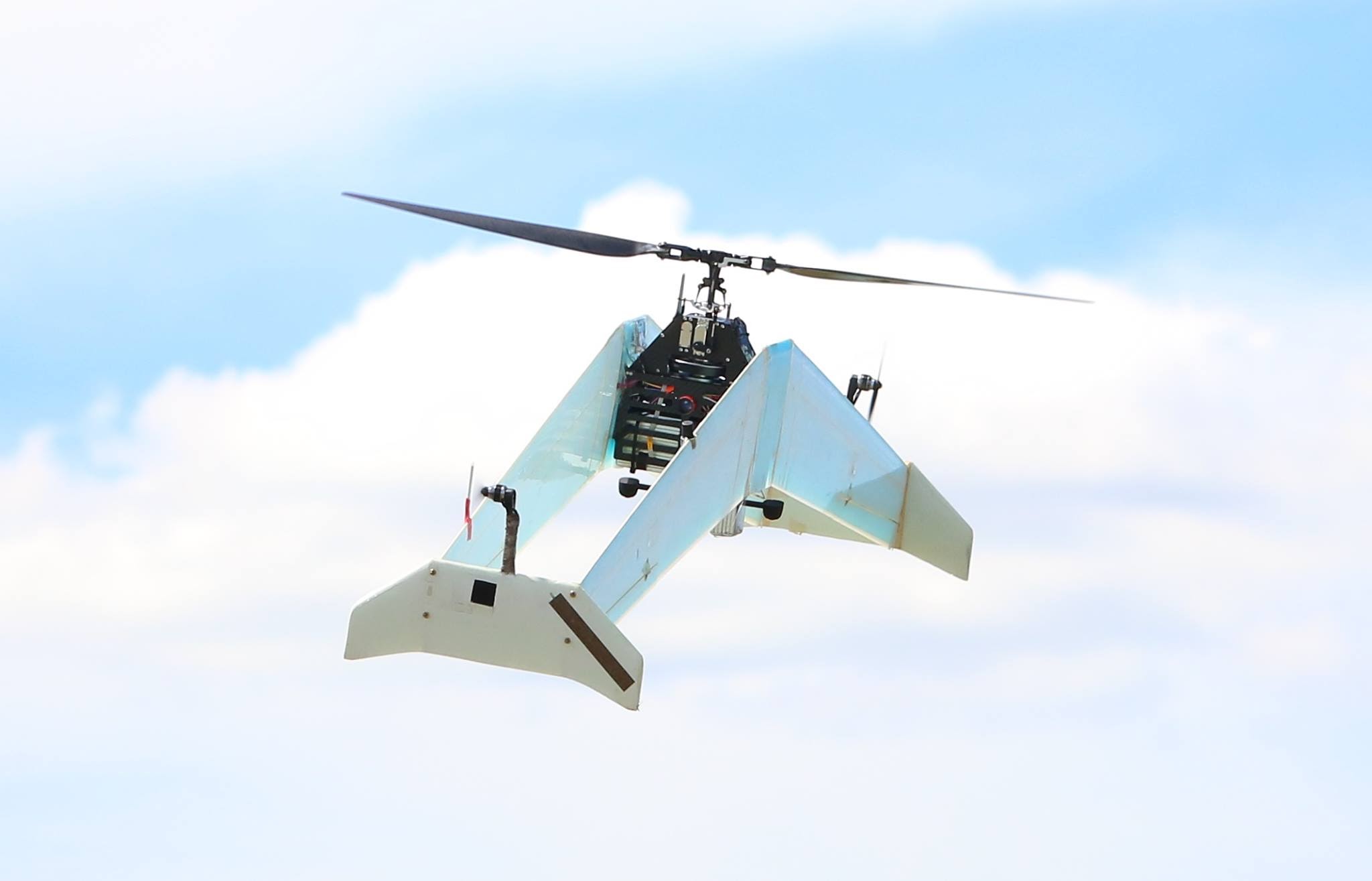}
\caption{Novel hybrid Unmanned Air Vehicle which combines a cyclic and collective pitch controlled main rotor with a biplane delta-wing and torque compensating tip rotors. The biplane concept adds structural rigidity and minimizes the lateral surface area to reduce the perturbations from turbulence during hover. The large main rotor allows efficient hovering flight while the cyclic control provides large control authority in hover.}
\label{figure:DelftacopterHover}
\end{figure}

\subsection{Medical Express Challenge}

A sample use-case for Vertical Take-Off and Landing (VTOL) aircraft with long-range capabilities is the Outback Medical Express UAV Challenge 2016.
The Outback UAV Challenge has a long history of creating realistic but very hard challenges for teams to improve the state of the art \cite{erdos2008uav,boura2011automated,erdos2013experimental}.

The 2016 edition of the Outback UAV Challenge was called \emph{Medical Express} and has set its competition goals to stimulate the development of aircraft with both hovering and long-range flight capabilities. The competition requires an unmanned vehicle to take-off  from a model airstrip in Dalby Australia and fly to a remote location $30$ km away which had often been inaccessible due to floods. At the location, a lost bush-walker must be located. The unmanned vehicle must then select a suitable landing location within $80$ meters from the found person, but for safety reasons may never come closer than $30$ meters to the person \cite{clothier2015structuring}. After an automatic vertical landing, medical assistance is delivered before flying back to base with a medical sample.

\subsection{Long-Distance VTOL}

\begin{figure}[hbt]
\centering
\def\svgwidth{10.4cm}

\begingroup%
  \makeatletter%
  \providecommand\color[2][]{%
    \errmessage{(Inkscape) Color is used for the text in Inkscape, but the package 'color.sty' is not loaded}%
    \renewcommand\color[2][]{}%
  }%
  \providecommand\transparent[1]{%
    \errmessage{(Inkscape) Transparency is used (non-zero) for the text in Inkscape, but the package 'transparent.sty' is not loaded}%
    \renewcommand\transparent[1]{}%
  }%
  \providecommand\rotatebox[2]{#2}%
  \ifx\svgwidth\undefined%
    \setlength{\unitlength}{150.23622047bp}%
    \ifx\svgscale\undefined%
      \relax%
    \else%
      \setlength{\unitlength}{\unitlength * \real{\svgscale}}%
    \fi%
  \else%
    \setlength{\unitlength}{\svgwidth}%
  \fi%
  \global\let\svgwidth\undefined%
  \global\let\svgscale\undefined%
  \makeatother%
  \begin{picture}(1,0.5754717)%
\iffinal
    \put(0,0){\includegraphics[width=\unitlength,page=1]{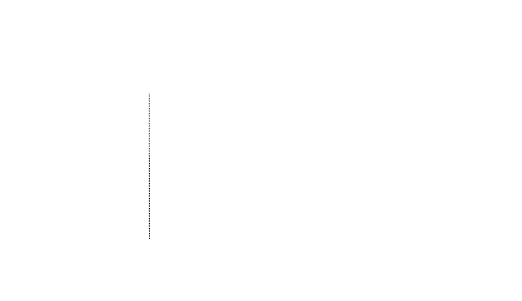}}%
    \put(0,0){\includegraphics[width=\unitlength,page=2]{drawings/transition.pdf}}%
    \put(0,0){\includegraphics[width=\unitlength,page=3]{drawings/transition.pdf}}%
    \put(0,0){\includegraphics[width=\unitlength,page=4]{drawings/transition.pdf}}%
\else
    \put(0,0){\includegraphics[width=\unitlength]{drawings/transition}}%
\fi
    \put(0.27699275,0.49818795){\color[rgb]{0,0,0}\makebox(0,0)[b]{\smash{lift from rotor}}}%
    \put(0.7417601,0.06568016){\color[rgb]{0,0,0}\makebox(0,0)[b]{\smash{forward}}}%
    \put(0.30793019,0.37001926){\color[rgb]{0,0,0}\makebox(0,0)[lb]{\smash{}}}%
    \put(0.29749765,0.0653103){\color[rgb]{0,0,0}\makebox(0,0)[b]{\smash{hover}}}%
    \put(0.71121445,0.52852886){\color[rgb]{0,0,0}\makebox(0,0)[b]{\smash{lift from wing}}}%
  \end{picture}%
\endgroup%

\caption{Lift generation in hover and forward flight.}
\label{figure:Transition}
\end{figure}

This paper proposes a novel UAV design (See Figure~\ref{figure:DelftacopterHover}) that combines efficient and high control authority hover with efficient long-range fast flight. While the design was optimized for the Outback Medical Express, it has applications far beyond. It features a large rotor for hovering with cyclic and collective pitch control combined with a biplane delta-shaped wing for efficient forward flight and all avionics and computer vision needed to turn it into a flying robot. During hover, all lift is provided by the main rotor and it uses tip-rotors and ailerons to compensate for the main rotor torque. During forward flight it pitches down almost $90$ degrees and transitions to a fixed-wing aircraft with a large propeller as illustrated in Figure~\ref{figure:Transition}.
Many drawbacks of this so-called tail-sitter when used for manned flight \cite{matsumoto2010hovering} don't apply to UAV, while its advantages remain.

Compared to coaxial counter rotating rotor designs, the chosen concept of a single rotor is much simpler and does not require axle in axle. In hover \delftacopter is basically a helicopter and in forward flight the motor rpm is reduced and the rotor blade pitch is increased to reach flight speeds of around $20$ to $25$ m/s to cover the required $60$ km in under an hour even in case of winds up to $25$ kt.

The flying wing of \delftacopter has the advantage of being simple and compact. It also yields advantages in the VTOL phases. Natural wind has a severe wind gradient close to the ground \cite{thornthwaite1943wind}.  When hovering, the top of the aircraft experiences a higher wind velocity than the lower part which calls for aircraft without a long tail. 

The choice for a biplane was made on three grounds:

\begin{itemize}
\item A single wing has more surface area exposed to the wind in VTOL mode than two wings behind each other. This diminishes the perturbations of take-off and landing in wind.
\item The two wings and fins at the tips yield a nice big footprint for stable ground handling.
\item A biplane configuration allows a higher range of angles of attack \cite{olson1976experimental} which is important in the transition from hover to forward flight and back.
\end{itemize}

\subsection{Outline}

The outline of the paper is as follows. First a propulsion system for both hover and forward flight is derived in Section~\ref{section:Propulsion}. Then the energy consumption (Section~\ref{section:Energy}) is addressed. Based on available propulsion and energy, the aerodynamic and structural design are detailed in Section~\ref{section:Airframe}. Wind tunnel measurements are analyzed in Section~\ref{section:Windtunnel}. The electrical design is explained in Section~\ref{section:Electrical}. Control of the \delftacopter is explained in Section~\ref{section:Rotormodel} and Section~\ref{section:Control}. The on-board computer vision follows in Section~\ref{section:Vision}. 
The flight is described in Section~\ref{section:Flight} and finally the Conclusions and Recommendations follow in Sections~\ref{section:Conclusions} and \ref{section:Recommendations}.


\section{Propulsion Design}\label{section:Propulsion}

The design of a propulsion system that is efficient in the wide range from fast forward flight down to stationary hovering flight is always a challenge. For \delftacopter, the propulsion is designed to be a compromise between efficient hover and efficient forward flight. This results in a rotor blade that is significantly different from rotors seen in conventional helicopters. 

The maximum efficiency for hover is obtained using a single large rotor \cite{bramwell2001bramwell}. Also for forward flight this is the most efficient \cite{anderson1999aircraft}. While in theory a single blade rotor is more efficient than a two blade rotor, in practice several implications still make the dual blade more practicable than a single blade.

The rotor blades are designed with significant twist, yielding a substantial performance increase over a flat rotor. The use of twist is possible thanks to the transitioning as---unlike in conventional helicopters---the rotor can always be kept in an axial flow regime.


\subsection{Propeller Design}

For an efficient hover the diameter has to be big enough to reach a reasonable figure of merit \cite{bramwell2001bramwell}. For forward flight where the power is significantly less than for hover the big diameter can be accepted when the rpm is reduced and the pitch is increased.

\begin{figure}[hbt]
\centering
\includematlab{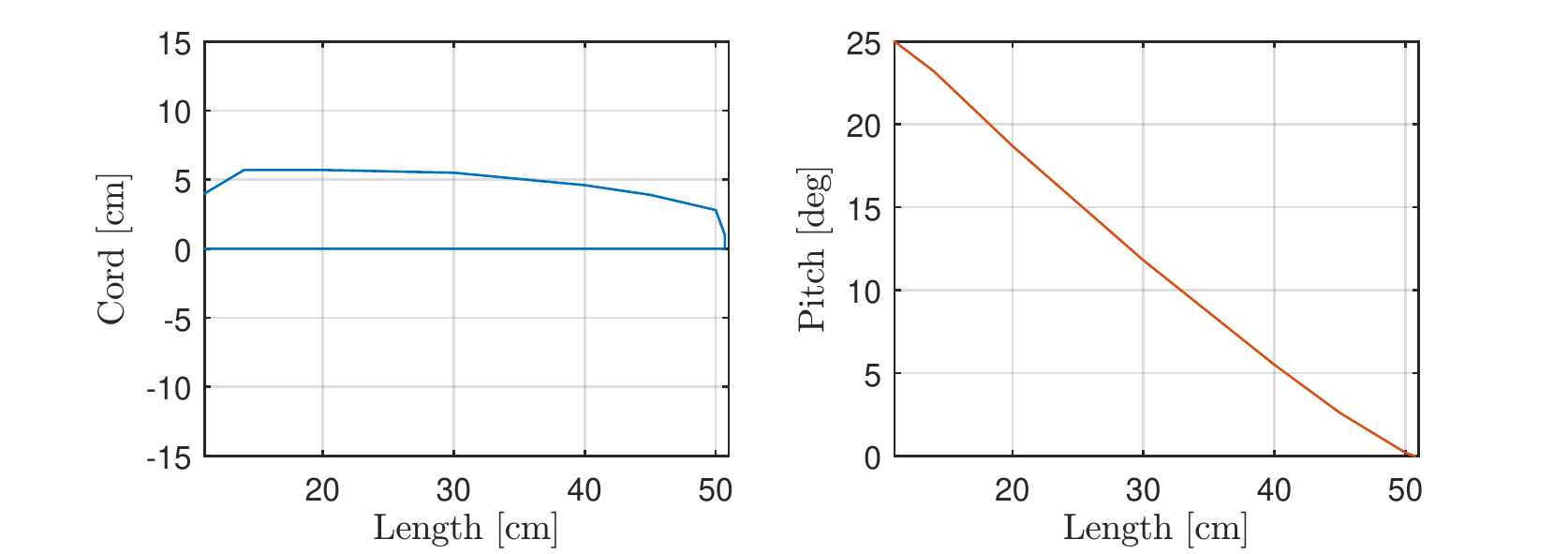}
\caption{Airfoil cord $c$ and pitch angle $\theta_0$ of the designed rotor blade.}
\label{figure:AirfoilSize}
\end{figure}

The design process was supported with the use of \emph{PropCalc 3.0} \cite{propcalc}\footnote{See \url{http://www.drivecalc.de/PropCalc/}}. A diameter of $1$ meter was selected as a between hover and forward flight requirements. A mild twist of $25$ degrees from root to tip was applied. For the airfoil the \emph{MA409} section was chosen being targeted at \emph{Reynolds} numbers of $Re_{0.7}=200.000$ and below.  The design point for hover was $1500$ rpm with $10^\circ$ tip pitch angle. For forward flight it was around $500$ rpm with $50^\circ$ tip pitch angle.

During the wind tunnel measurements and  flight tests the drag of all external appendages was found to be higher than estimated in the concept phase, which gave us the need to lower the pitch and increase the rpm. This also increased the responsiveness needed to climb more quickly when required. The new cruise condition was around $30^\circ$ tip pitch angle with $910$ rpm. In cases where responsiveness or excess thrust are required, $23^\circ$ pitch with $1140$ rpm was utilized.

\begin{figure}[phbt]
\centering
\includematlab{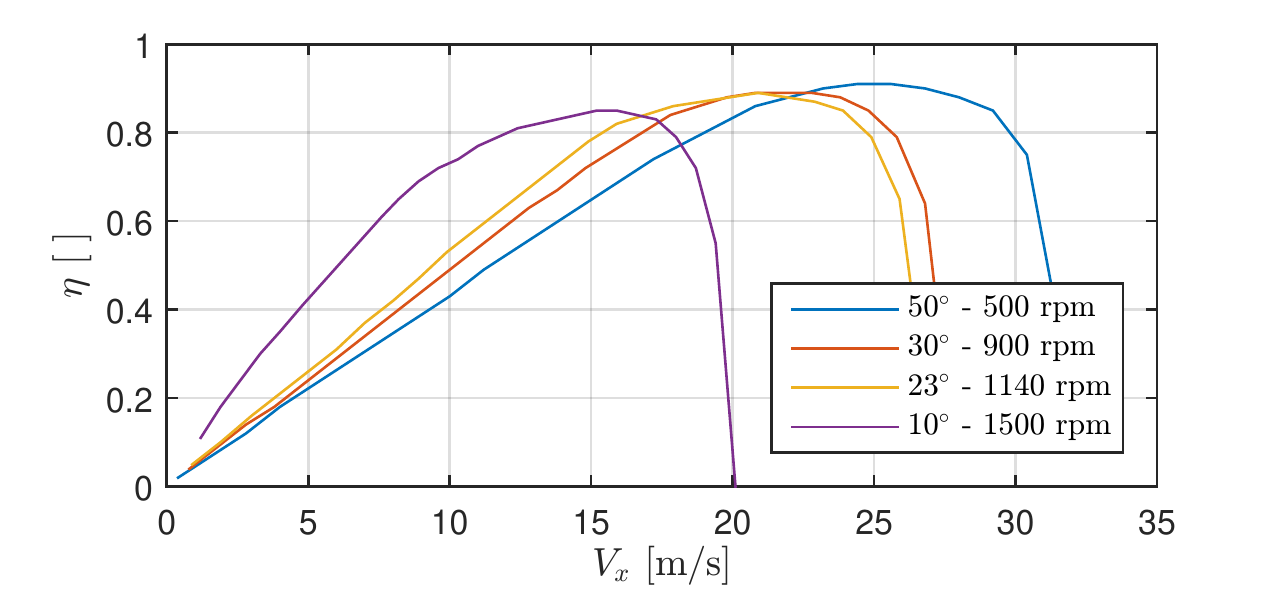}
\caption{Propulsion efficiency at selected pitch angles and rpm.}
\label{figure:PropEffrpm}
\end{figure}

\begin{figure}[phbt]
\centering
\includematlab{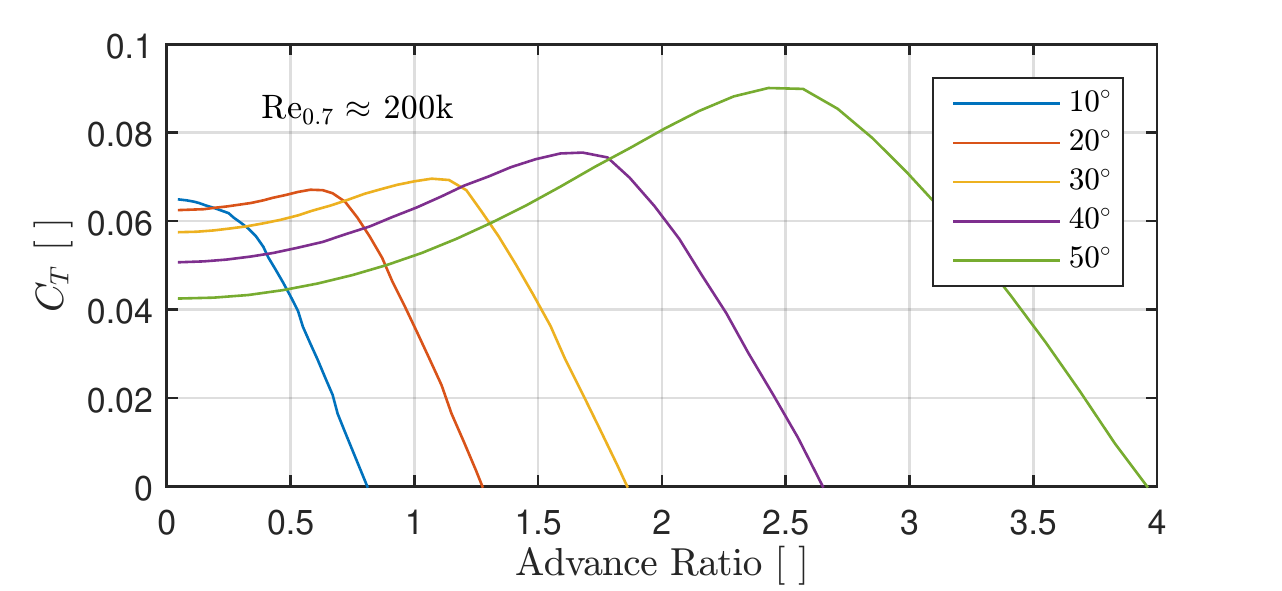}
\caption{Propulsion thrust coefficient.}
\label{figure:PropCT}
\end{figure}

\begin{figure}[phbt]
\centering
\includematlab{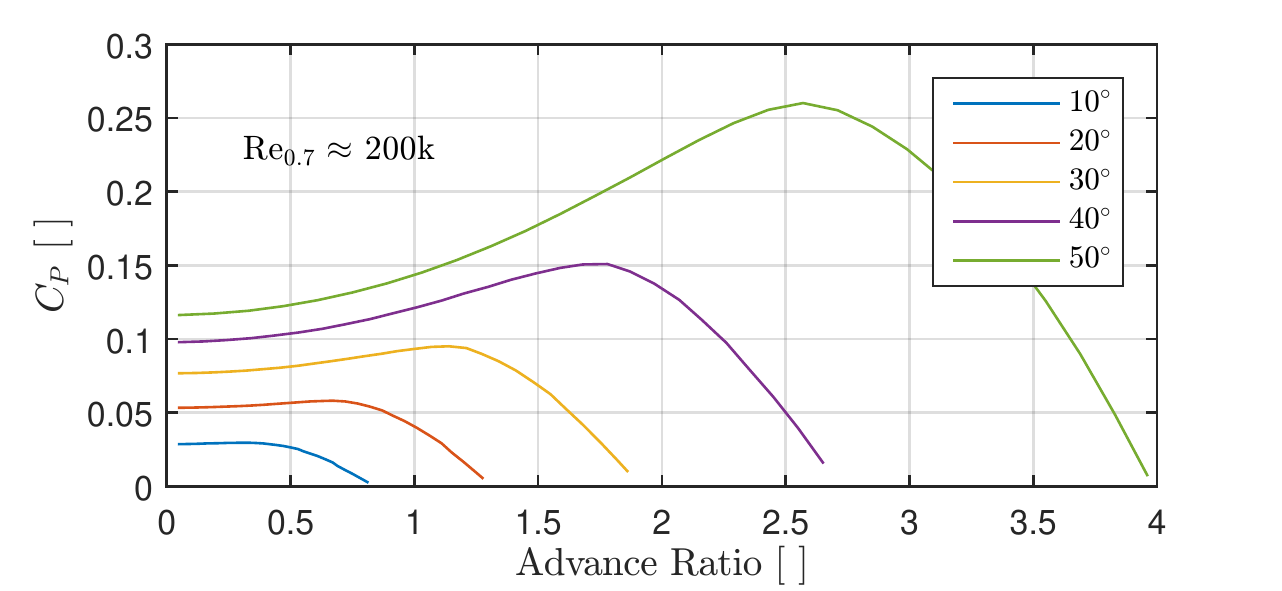}
\caption{Propulsion power coefficient.}
\label{figure:PropCP}
\end{figure}

\begin{figure}[phbt]
\centering
\includematlab{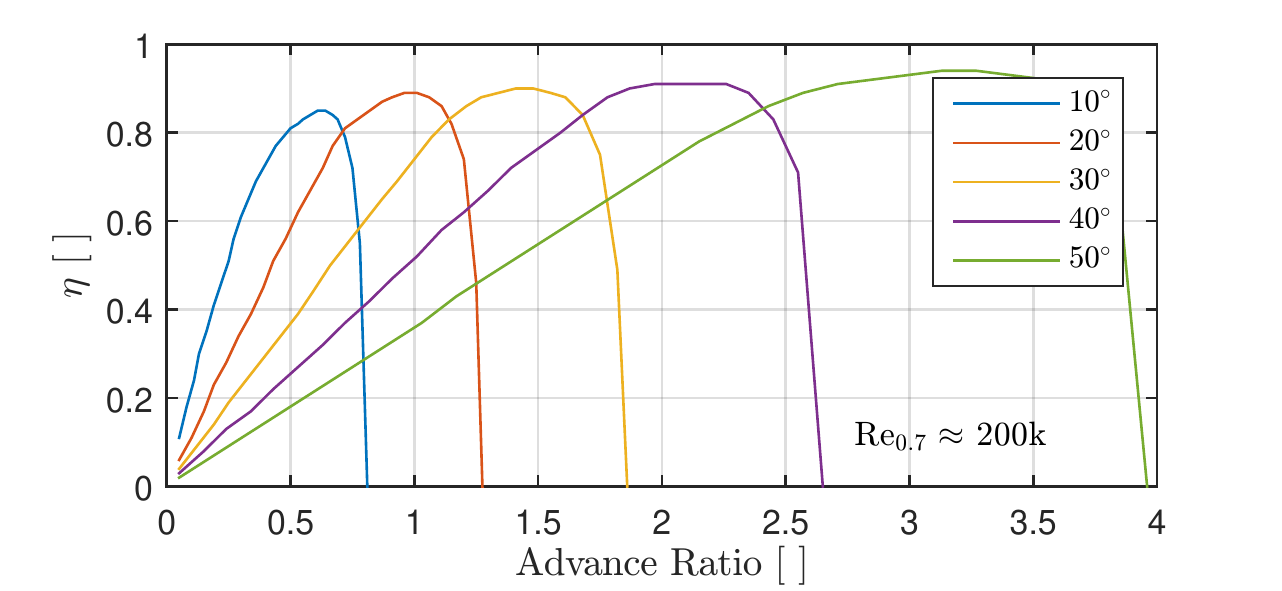}
\caption{Propulsion efficiency.}
\label{figure:PropEff}
\end{figure}

The resulting propeller is shown in Figure~\ref{figure:AirfoilSize} and the size is given in Table~\ref{table:AirfoilSize}. Characteristics are given in Figures~\ref{figure:PropCT},\ref{figure:PropCP},\ref{figure:PropEff},\ref{figure:PropFrpm},\ref{figure:PropPrpm} and~\ref{figure:PropEffrpm}.

\begin{table}[hbt]
	\caption{Airfoil cord $c$ and pitch angle $\theta_0$.}
	\label{table:AirfoilSize}
	\begin{center}
		\begin{tabular}{|r||l|l|}
			\hline
x [cm] & $c$ [cm] &  $\theta_0$ \\ \hline
\hline
11 &  4   & 25.0  \\ \hline
14 &  5.7 & 23.2 \\ \hline
20 &  5.7 & 18.7  \\ \hline
30 & 5.5  & 11.8  \\ \hline
40 &  4.6 & 5.5 \\ \hline
45 &  3.9 & 2.6 \\ \hline
50 &  2.8 & 0.2 \\ \hline
50.7 &1.0 & 0 \\
			\hline
		\end{tabular}
  \end{center}
\end{table}

\subsection{Motor}

Once the propeller design was shown to yield good efficiency in both flight regimes, a corresponding motor was chosen capable of delivering the required torque and power. 
A $105kV$ direct-drive sensor-less brush-less direct current (BLDC) motor was selected.

\subsection{Silent}

\delftacopter produces noise equivalent to a medium sized quadrotor when hovering, with most noise originating from the tip props.
But when in forward flight, \delftacopter becomes very silent.
During the transition from hover to forward flight, most of the noise disappears when the $6$ x $4.5$ inch tip propellers are turned off.
This reduced noise production is considered to be a significant benefit of using one large efficient rotor with high pitch, low RPM and a direct drive motor.

\begin{figure}[phbt]
\centering
\includematlab{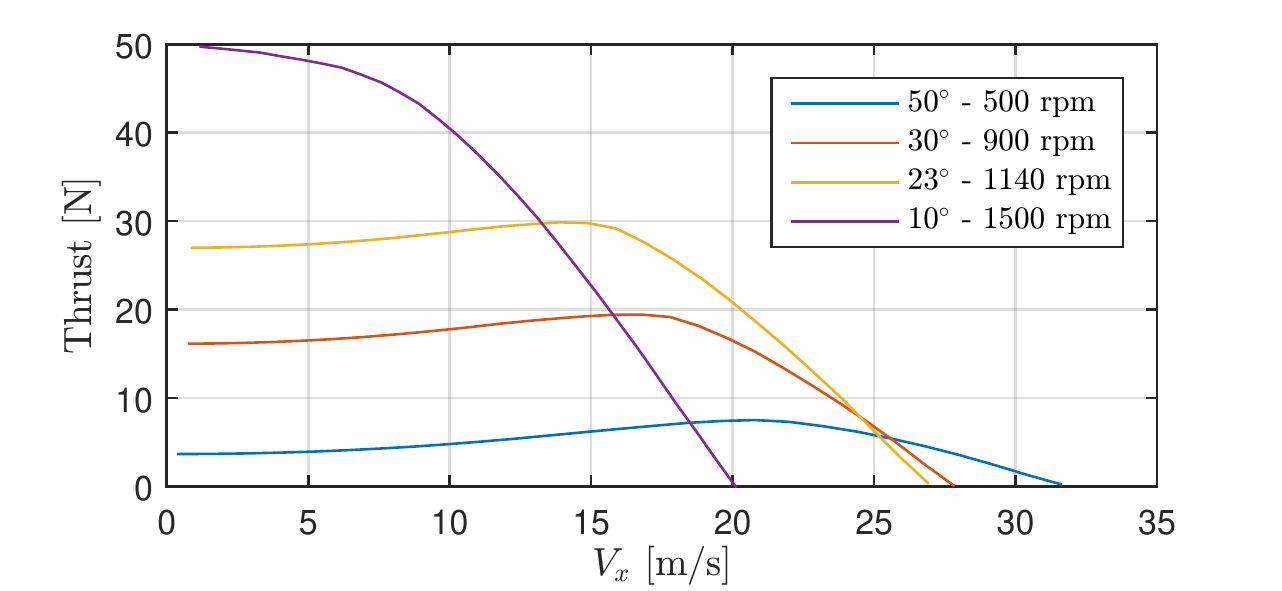}
\caption{Propulsion thrust at selected pitch angles and rpm. Higher thrust for hover or forward acceleration can only be obtained at higher rpm.}
\label{figure:PropFrpm}
\end{figure}

\begin{figure}[phbt]
\centering
\includematlab{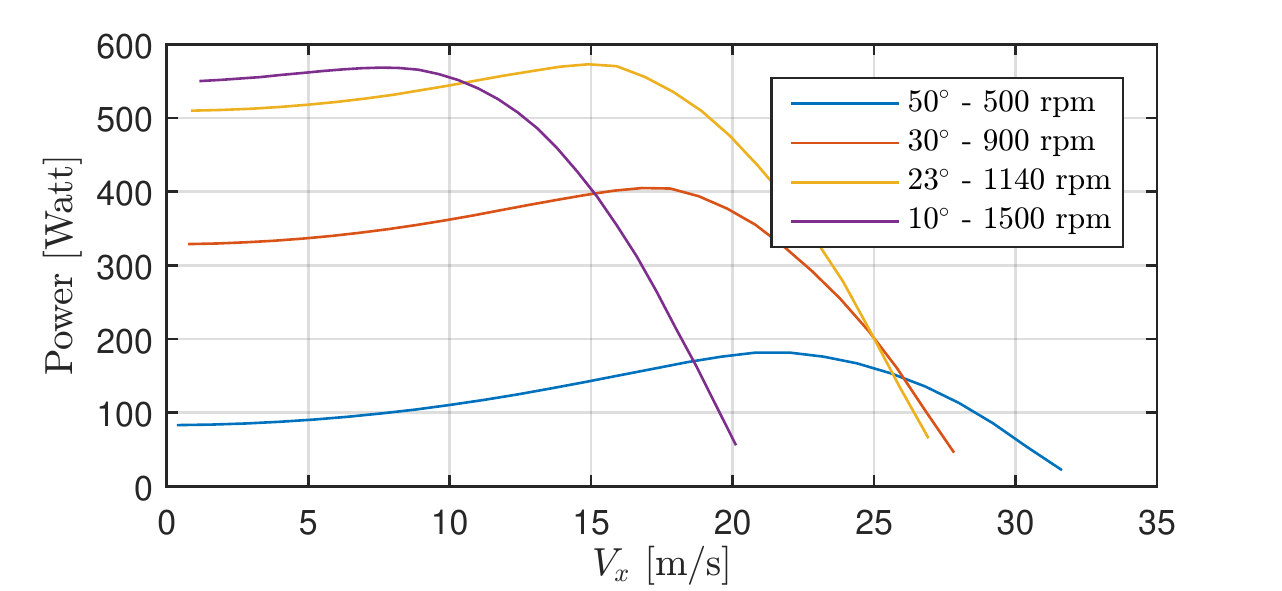}
\caption{Propulsion power at selected pitch angles and rpm.}
\label{figure:PropPrpm}
\end{figure}

\section{Energy}\label{section:Energy}

\begin{figure}[hbt]
\centering
\includematlab{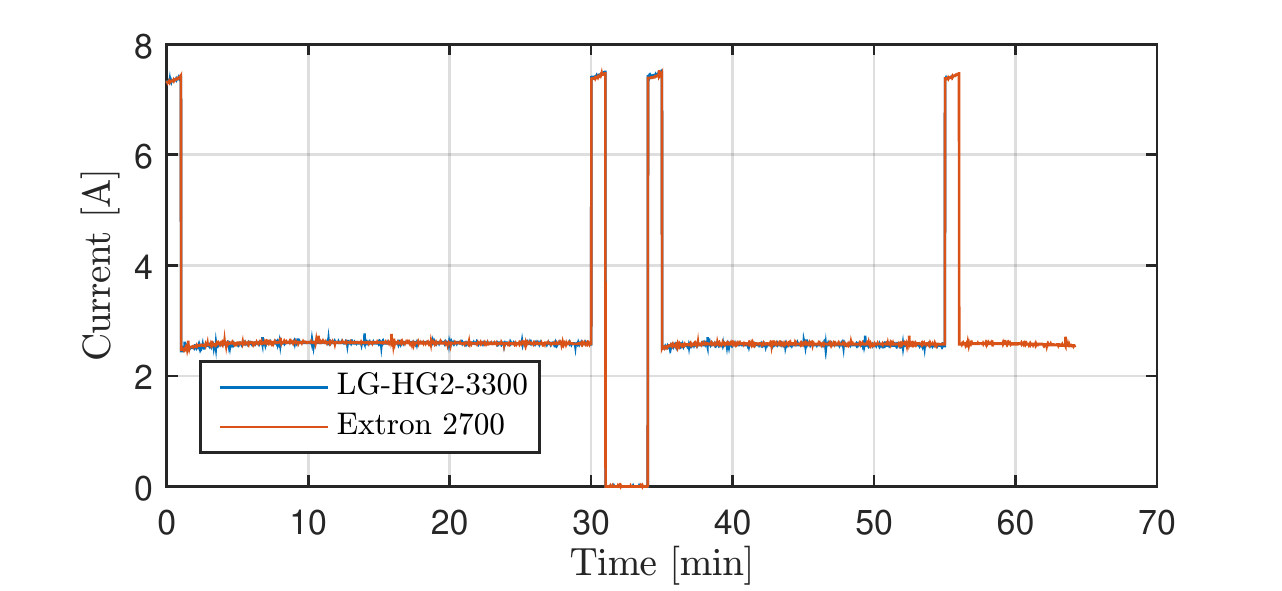}
\caption{Test discharge current profile. The profile simulates a 1 minute hovering take-off, followed by
an efficient $29$ min forward cruising flight, a $1$ min hovering landing, $3$ min of waiting time and the same return flight. }
\label{figure:EnergyDischargeAmp}
\end{figure}

An energy profile was derived from the mission requirements. It consists of a higher load phase during vertical take-off, followed by an endurance phase at lower load and another high load phase during the landing. For the competition, after a short down time, there is also a return flight with the same profile (See Figure~\ref{figure:EnergyDischargeAmp}).

Several types of electrical energy are available. Common battery technologies
for electric UAV are Lithium-Polymer and Lithium-Ion batteries. Even higher energy densities can be achieved using fuel cells \cite{larminie2003fuel}. But because of the short mission time of less than 1 hour and high flight speed involved in the competition, the depletion rate is also important. 

\begin{figure}[hbt]
\centering
\includematlab{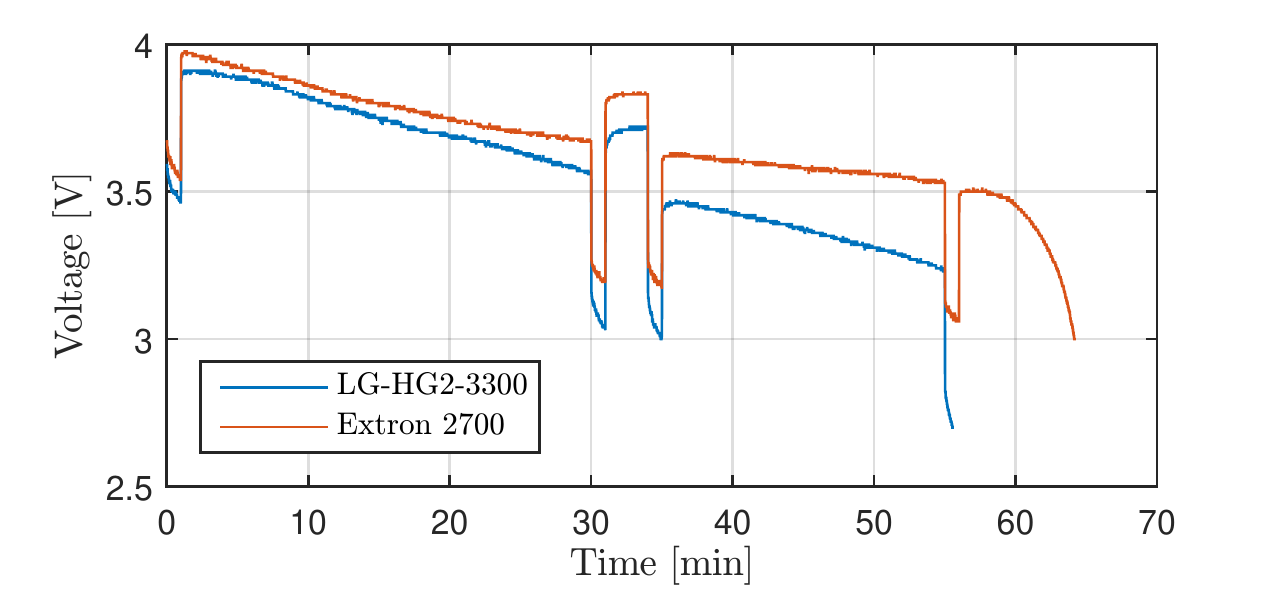}
\caption{Discharge voltage in function of time for a Lithium-Polymer versus Lithium-Ion battery subjected to the mission load profile.}
\label{figure:EnergyDischargeVolt}
\end{figure}

No fuel cells could be found within the weight budget and power rating. 
To make the choice between the more energy dense Lithium-Ion and higher current rated Lithium-Polymer types a test setup was built.
The \emph{LG-HG2-3300} Lithium-Ion battery was trade off against the \emph{Extron 2700} Lithium Polymer battery in a test setup which submitted the cells to the competition load. Figure~\ref{figure:EnergyDischargeVolt} shows the test results in which large differences can be observed. While the Lithium-Ion cells contain $22\%$ more energy under slow discharge, under high load they delivered less power and in the end  could not deliver the needed power needed for the final landing. Finally the Lithium-Polymer cells were selected for \delftacopter.



\section{Airframe Design}\label{section:Airframe}

Given the propulsion system, energy package and rotor-head mechanics, a fixedwing airframe was designed. The airframe needs to generate lift during fast forward flight with little drag, but at the same time it must also accommodate all the systems of the flying robot. Finally it must provide structural stability for the airframe to land as a rotorcraft.

\begin{figure}[hbt]
\centering
\includegraphics[width=10.4cm]{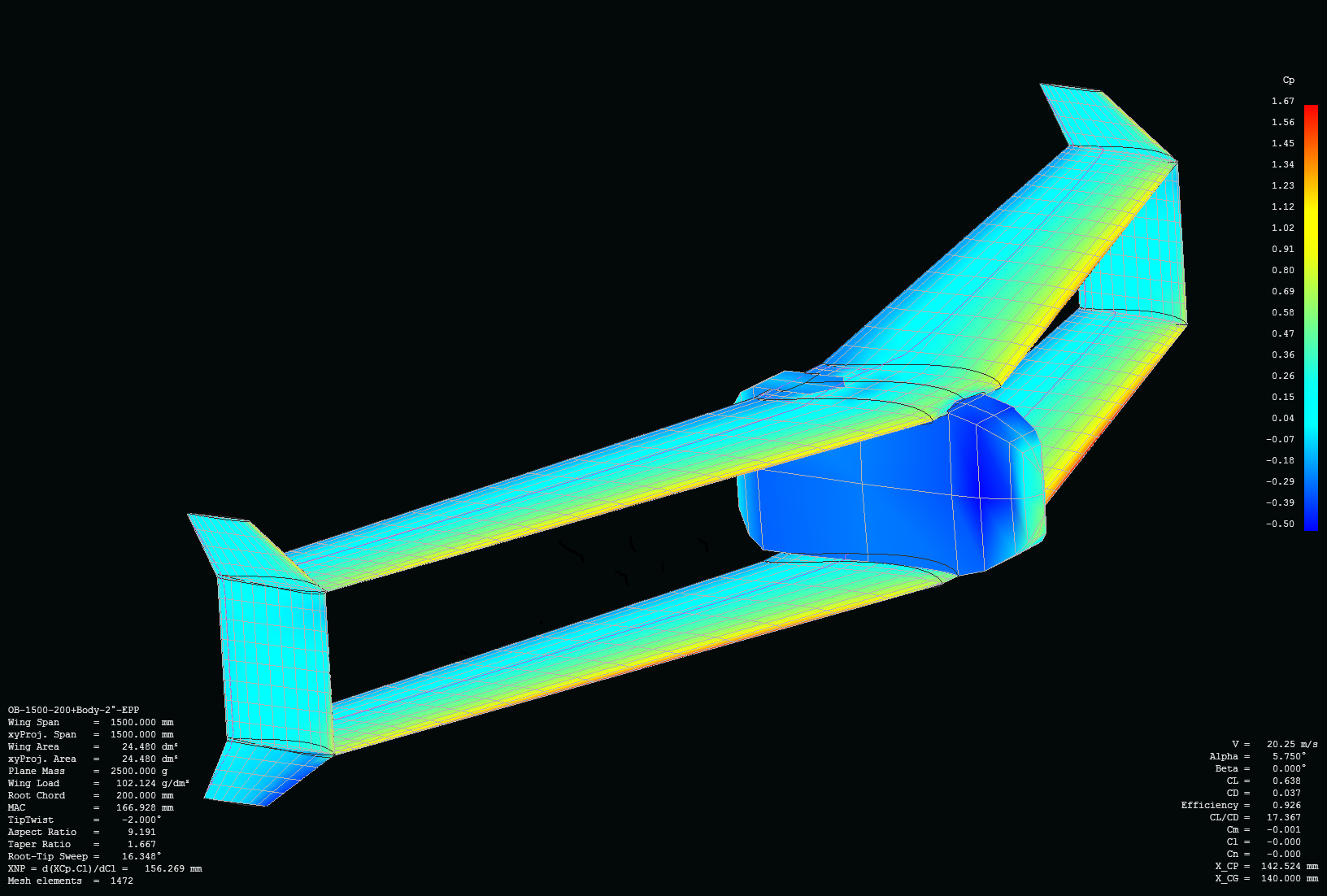}
\caption{Screen-shot from XFLR drag computations. The biplane wings and wing tips as well as the fuselage are modeled. The color shadings reflects the pressure coefficient $C_P$.}
\label{figure:AeroXflr}
\end{figure}

\subsection{Structural}

When \delftacopter is in hover, the wing acts as a landing gear but also makes the helicopter more sensitive to lateral gusts. To address both problems at once, a biplane wing was selected. This has the advantage that the total lateral surface area in hover is significantly reduced. This means its size and corresponding moments are reduced, while the two wings provide a stable rectangular basis for landing.

\subsection{Aerodynamic Design}

\begin{figure}[hbt]
\centering
\includematlab{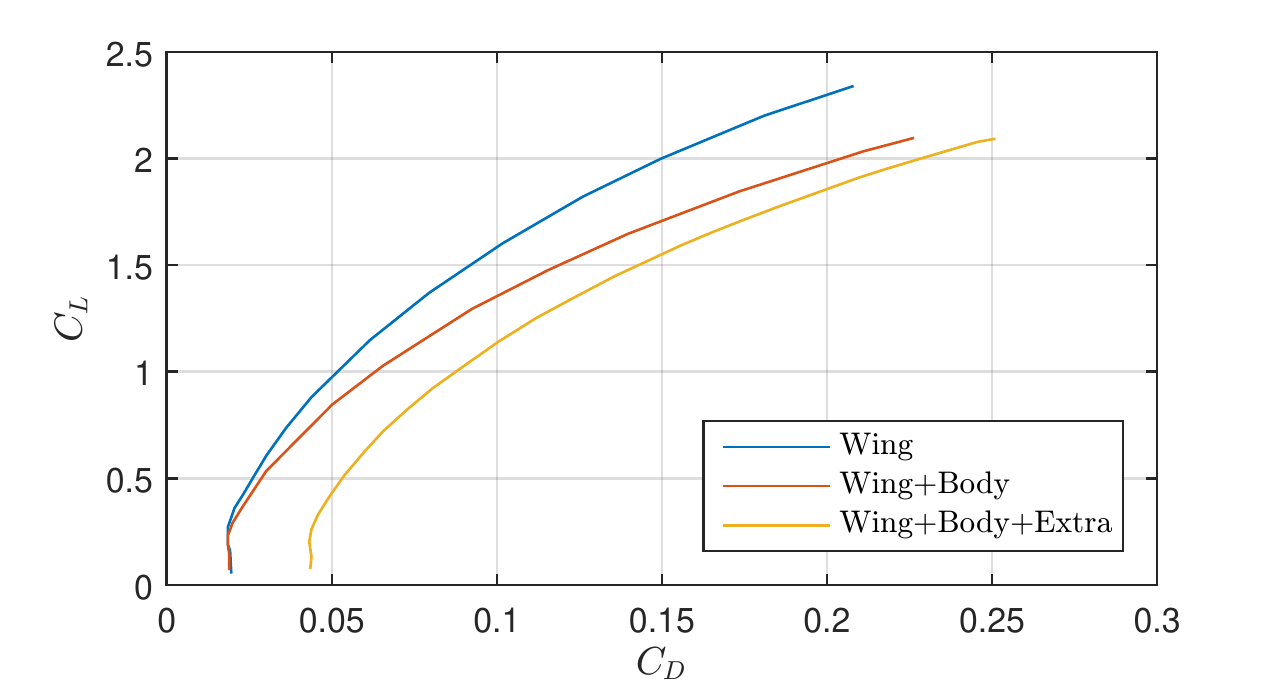}
\caption{Lift-Drag polar computation using XFLR for a $4.5$ kg \delftacopter with a c.g. at $x=140.0$ mm from the central section leading edge. The figures show computed polars for the wings only, the wings with ideal fuselage and an estimate for the total vehicle including drag from rotor head and all protruding items like antennas.}
\label{figure:AEROClCd}
\end{figure}

\begin{figure}[hbt]
\centering
\includematlab{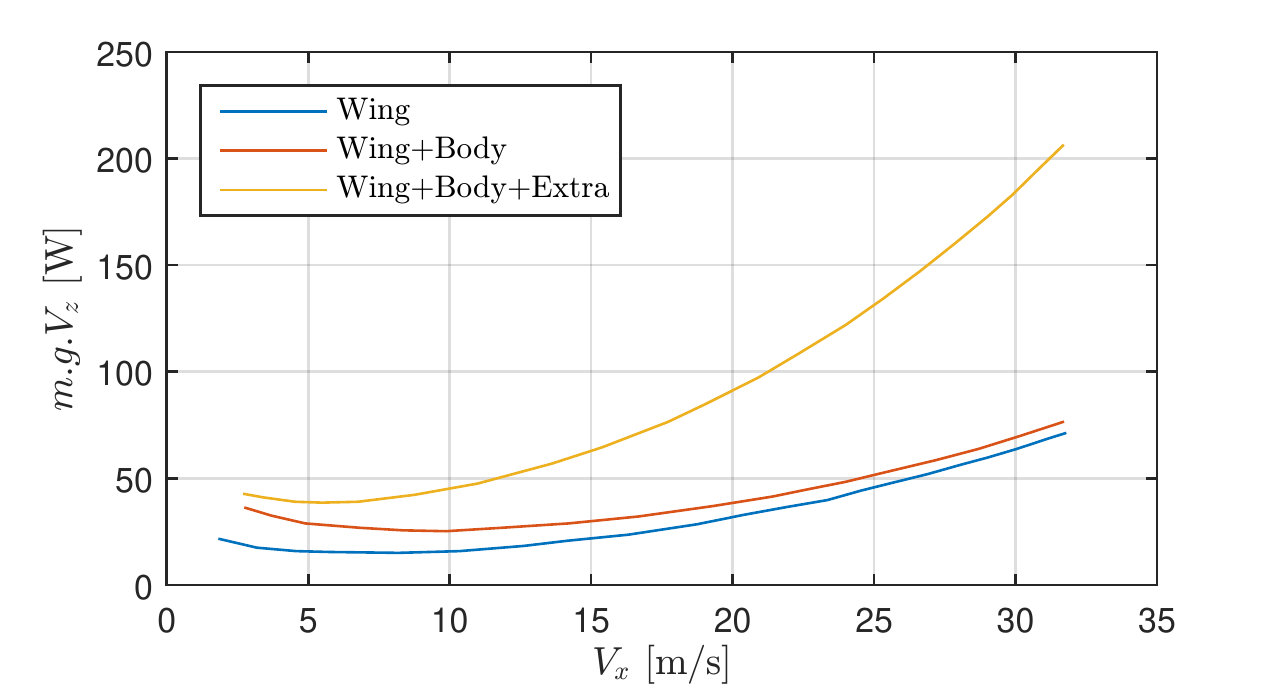}
\caption{Required aerodynamic power in function of forward speed computation using XFLR. It is clearly visible that especially at higher speeds a lot can still be gained by reducing the parasitic drag of \delftacopter.}
\label{figure:AEROVxW}
\end{figure}

A delta-shaped auto-stable flying wing concept was selected. This removes the need for a vertical stabilizer and fits well behind the main rotor. Lift and Drag computations were performed using XFLR \cite{drela2012xflr}. Figure~\ref{figure:AeroXflr} shows a view from the 3D model.

Figure~\ref{figure:AEROClCd} shows the insertion of the fuselage has almost no influence on $C_d$ at $C_l=0$ due to the inviscid calculation. At higher angles of attack the influence is significant. The drag due to the non-streamlined fuselage, the rotorhead, motor cooling and all protrusions like head antennae etc is added as an extra term. This is taken as a $C_{d_0} = 0.012$ based on the total wing area to lead to a more realistic drag. 

Nevertheless, comparison between Figures~\ref{figure:PropPrpm} and \ref{figure:AEROVxW} shows that even with the extra drag, the power required to fly forward at $25$ m/s is significantly less than the power needed to hover or fly using the rotor alone thanks to the delta-wing.



\section{Wind Tunnel Analysis}\label{section:Windtunnel}

\begin{figure}[hbt]
\centering
\includegraphics[width=10.4cm]{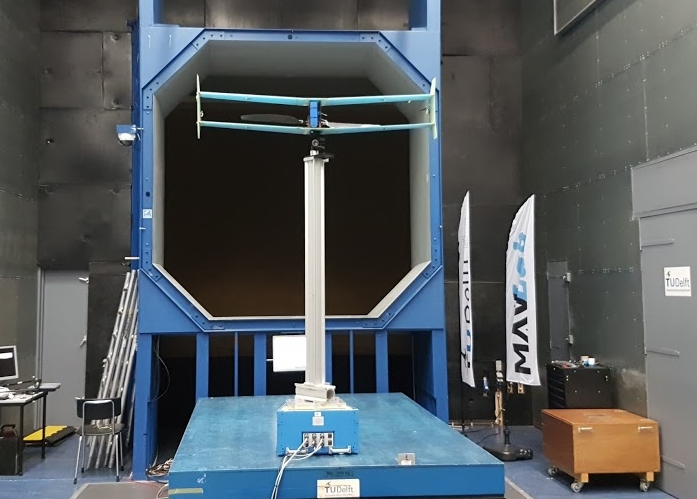}
\caption{\delftacopter in the Open Jet Windtunnel of TUDelft. The tunnel outlet measures $2.85$ m by $2.85$ m and can reach $30$ m/s wind speeds.}
\label{figure:Windtunnel}
\end{figure}

In a typical scenario, \delftacopter would spend most of its time in forward flight.
Therefore, optimizing the energy efficiency in forward flight is key to improving the range of the vehicle.
While computations in Section~\ref{section:Propulsion} predicted an efficiency increase in forward flight with lower
rpm, this could not easily be seen during flight tests. A possible explanation could be that the motor is less
efficient at low RPM, canceling the performance gain from the propeller. To verify this, windtunnel measurements were performed.

In order to assess the propulsive efficiency and to find the optimal propulsion settings, a wind tunnel experiment was performed.
The Open Jet Facility at Delft University of Technology was used for this experiment.
The vehicle was placed in the middle of the $2.85$ m by $2.85$ m wind tunnel outlet, with zero angle of attack.
The \delftacopter was rigidly attached to a pole, which was mounted on a force-moment balance below the wind tunnel outlet as shown in Figure~\ref{figure:Windtunnel}.

\begin{figure}[hbt]
\centering
\includematlab{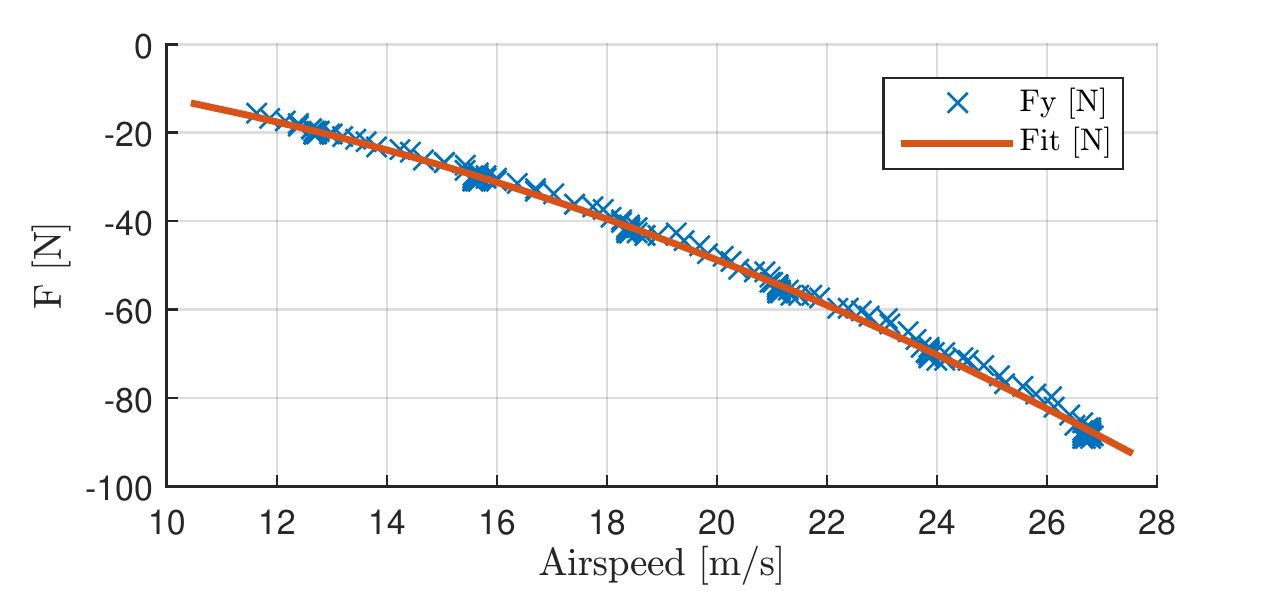}
\caption{Drag of the pole in function of airspeed.}
\label{figure:Wind:Poledrag}
\end{figure}

First the drag of the pole and attachment without the \delftacopter was measured. Figure~\ref{figure:Wind:Poledrag} shows the resulting fit which finds a drag of $D = \frac{\rho}{2} V^2 \cdot 0.195$.
Then \delftacopter is mounted on the pole in the middle of the open jet wind tunnel. Measurements are taken at several representative airspeeds, namely at $15$, $19$, $24$ and $27$ m/s. For each airspeed, \delftacopter parameters are measured through a range of main rotor collective pitch angles and power settings. The settings are selected manually such that no rpm, current or motor temperature limitation is breached. The rotor is turning at all times as soon as the wind tunnel is blowing, and windmills even when no power is applied.

\begin{figure}[hbt]
\centering
\includematlab{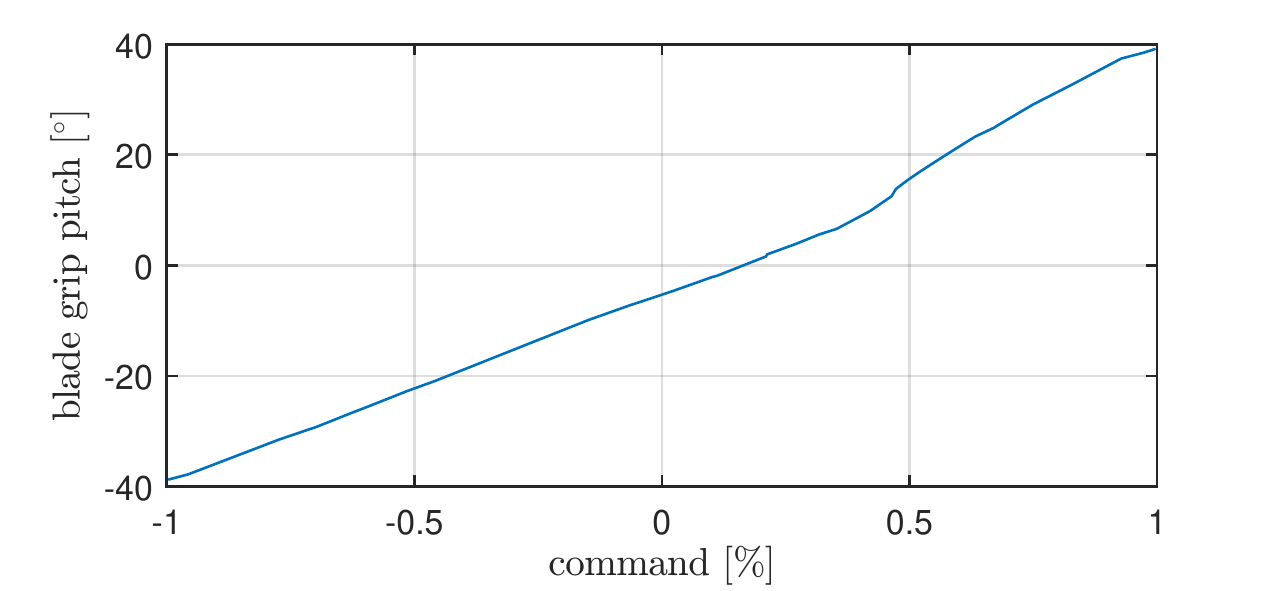}
\caption{The non-linearity of the collective pitch angle of the main rotor in function of the scaled servo command.}
\label{figure:Wind:PitchNonLin}
\end{figure}

One extra lab measurement is made to convert the servo pitch commands into an actual collective pitch angle. The non-linearity of the rotor-head  linkages is non-trivial as seen in Figure~\ref{figure:Wind:PitchNonLin}. A close-up photo of the rotor-head can be found in Section~\ref{section:Control}  Figure~\ref{figure:HeadCloseUp}.

\begin{figure}[hbtp]
\centering
\makebox[\textwidth][c]{\includematlab{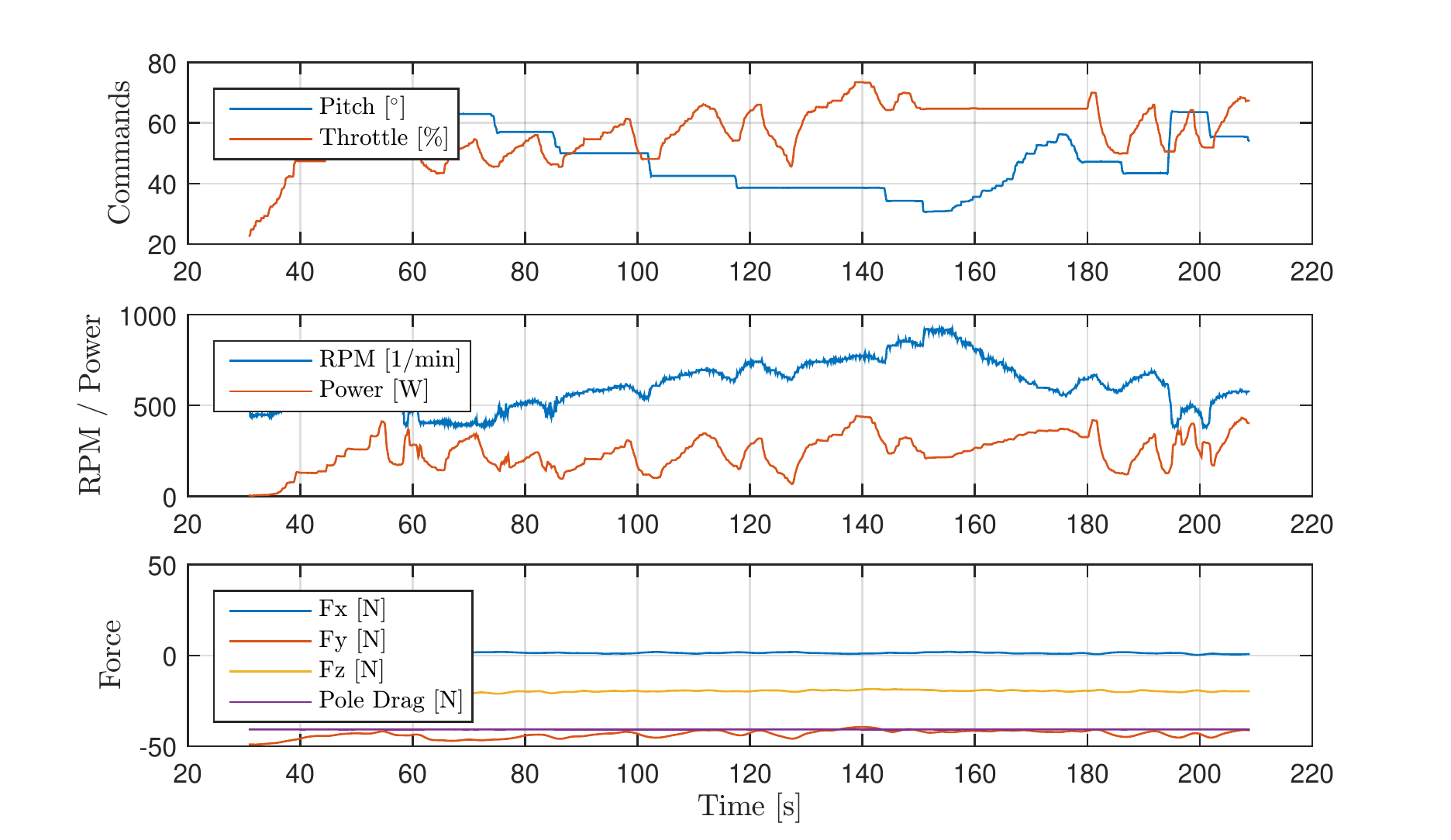}}
\caption{RAW windtunnel data and on-board measurements for the tunnel setting at $19$ m/s. For every pitch setting all acceptable throttle settings are visited and the effect on power use and forward thrust is measured.}
\label{figure:Wind:Rawdata19}
\end{figure}

During the windtunnel runs, all on-board data like IMU and airspeed are logged. Of special interest are the motor current and rpm together with the throttle setting and collective pitch setting. In parallel the windtunnel system logged all forces and moments on the balance and the windtunnel calibrated airspeed (See Figure~\ref{figure:Wind:Rawdata19}).
From the data it was expected to find a clear sweet spot, namely a throttle versus pitch setting where better efficiency could be obtained.

\begin{figure}[hbtp]
\centering
\includematlab{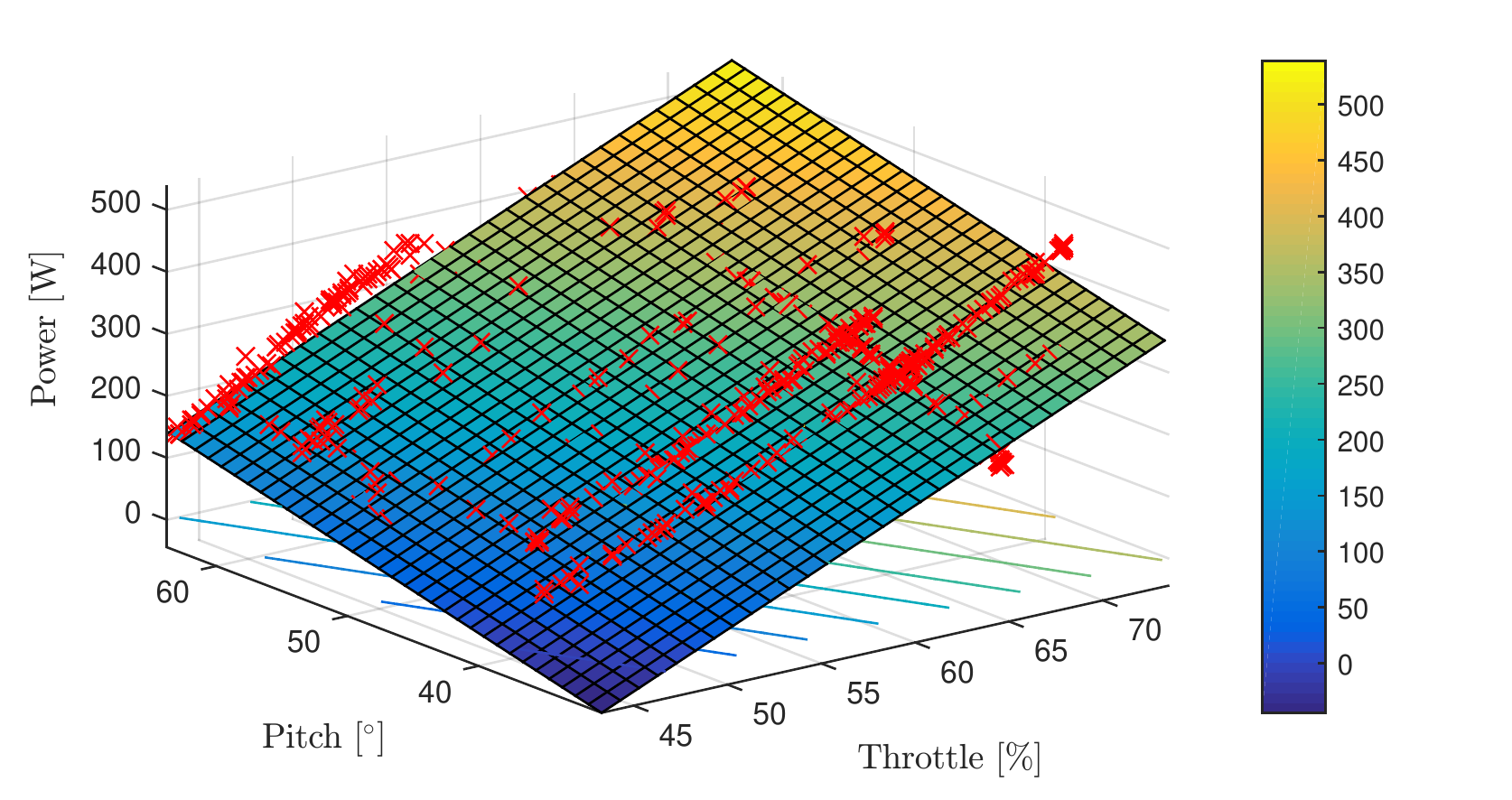}
\caption{Measured power in function of pitch and throttle and planar fit showing rpm and pitch can be exchanged while keeping the same used power.}
\label{figure:Wind:PitchThrottle19}
\end{figure}

However, it was found that power and rpm wise, pitch and throttle can be exchanged without significant difference in power efficiency. Figure~\ref{figure:Wind:PitchThrottle19} shows a planar fit predicting the stationary used power based on the pitch and throttle input. Identical results are obtained at other windtunnel velocities. The plane fits the data very well with most off-plane points corresponding to temporary changes in power setting. When in Figure~\ref{figure:Wind:PitchThrottle19} contour lines of the fit are followed from left to right, one finds settings for pitch and throttle that consume the same amount of power but result in a different rpm.

\smallskip

Several other interesting observations can be made from the windtunnel data. The force graph in Figure~\ref{figure:Wind:Rawdata19} for instance shows that \delftacopter does not have a very large spare thrust in fast forward flight. This looks significantly less than the values found in Figure~\ref{figure:PropFrpm} because of electrical inefficiencies and because the windtunnel test did not include points overloading the motor.
Finally it can be seen that the highest thrust $T=F_y$ is obtained at higher rpm values.

Overall it can be concluded that \delftacopter can exchange rpm and pitch without very significant change in efficiency but that higher thrust can be achieved with higher rpm and higher maximum velocity can only be achieved with higher pitch settings.


\section{Electronic Design}\label{section:Electrical}

To comply with the strict requirements of the Outback Medical Challenge \cite{clothier2015structuring} and be allowed to fly beyond visual line of sight (BVLOS) missions at up to $30$ km distance, a custom electronic design was required. 
It consists of two independently powered circuits.

\begin{figure}[hbtp]
\centering
\includegraphics[width=1.0\linewidth]{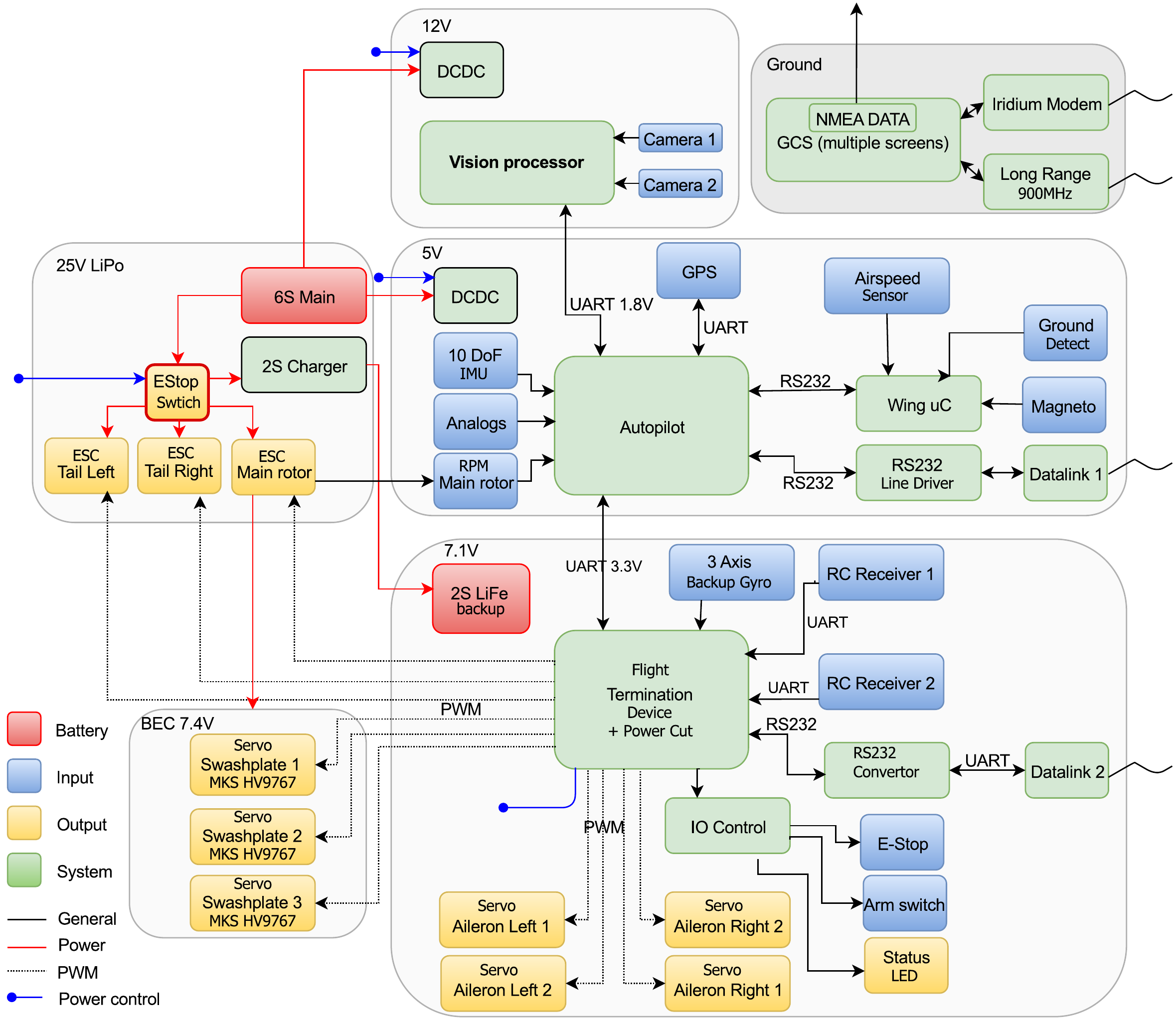}
\caption{Schematic overview of \delftacopter electronics}\label{figure:Electronics}
\end{figure}

As seen in Figure~\ref{figure:Electronics}, the first part is called flight termination device.
This part has all the safety critical functions like driving actuators but also geo-fencing and long-range kill switches, motor un-powering and arming.

All navigation and control functions together with the flight plan logic are in the other part called autopilot.
Both parts are modification of the Paparazzi-UAV \cite{brisset2006paparazzi} Lisa-MX autopilot \cite{gati2013open}.

\begin{figure}[hbtp]
\centering
\includegraphics[width=10.4cm]{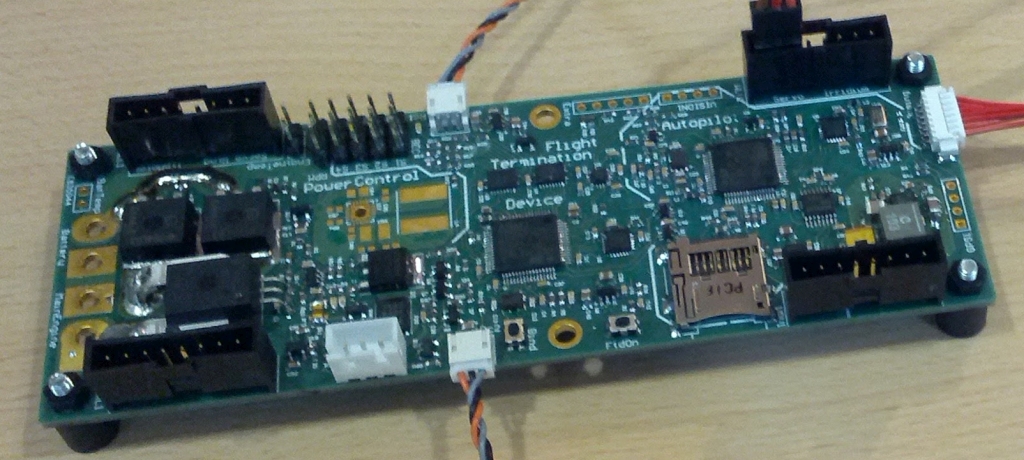}
\caption{All central electrical functions of \delftacopter are integrated into a single board PCB for minimal weight and minimal interconnection failure. The four corner connectors lead to the systems in all four wings. From left to right the board contains power, flight termination and autopilot.}\label{figure:ElectronicsPhoto}
\end{figure}

Because of the number of extra functions and boards, like SD-card logging, master power cut-off, line drivers to modems in wings, power converters, current voltage and temperature sensors, the design started to grow larger.
To minimize interconnection failures and minimize the total weight, a custom PCB was designed with all needed functions which is shown in Figure~\ref{figure:ElectronicsPhoto}.

\section{Rotorhead Dynamics}\label{section:Rotormodel}

Rotorcraft dynamics have been well studied  for many years \cite{bramwell2001bramwell,prouty1995helicopter,wagtendonk1996principles,johnson1980comprehensive,stepniewski1979rotary,gavrilets2015dynamic,shim1998comprehensive,padfield2008helicopter}. 
But the properties of the light efficient rotor on a large heavy fuselage found in \delftacopter are significantly different from what is seen in conventional helicopter control.

\begin{figure}[hbtp]
\centering
\includegraphics[width=10cm]{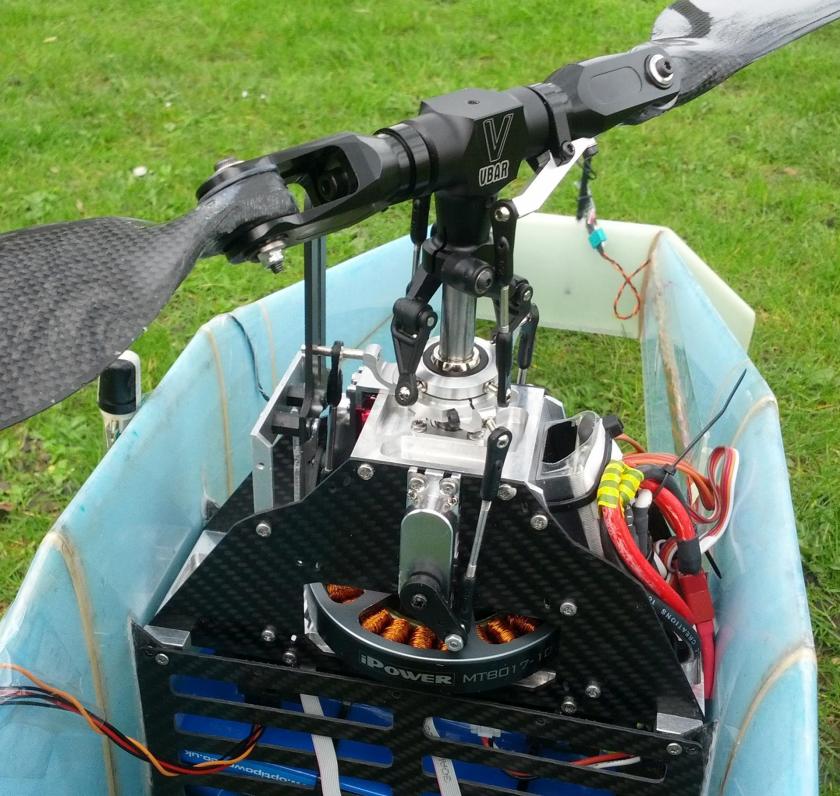}
\caption{Close up of the rotor head of \delftacopter. The swash-plate has three gripping points at $120$ degrees from each other. Collective pitch has double the pitch range from conventional helicopters. Collective pitch can reach from $-40$ to $40$ degrees. The self laminated blades with \emph{MA409} section have high camber and high lift coefficients and $25$ degrees of blade twist from root to tip. Hovering flight is performed at a designed tip angle of attack of around $10^\circ$ with $1500$ rpm. In forward flight the tip angle of attack can change up to $50^\circ$ at $500$ rpm. The root angle of attack is then about $75^\circ$. The blade twist is made possible because the flow is always axial as \delftacopter transitions. This allows the rotor to be efficient from hover to fast forward flight.}
\label{figure:HeadCloseUp}
\end{figure}

\begin{figure}[hbt]
\centering
\includematlab{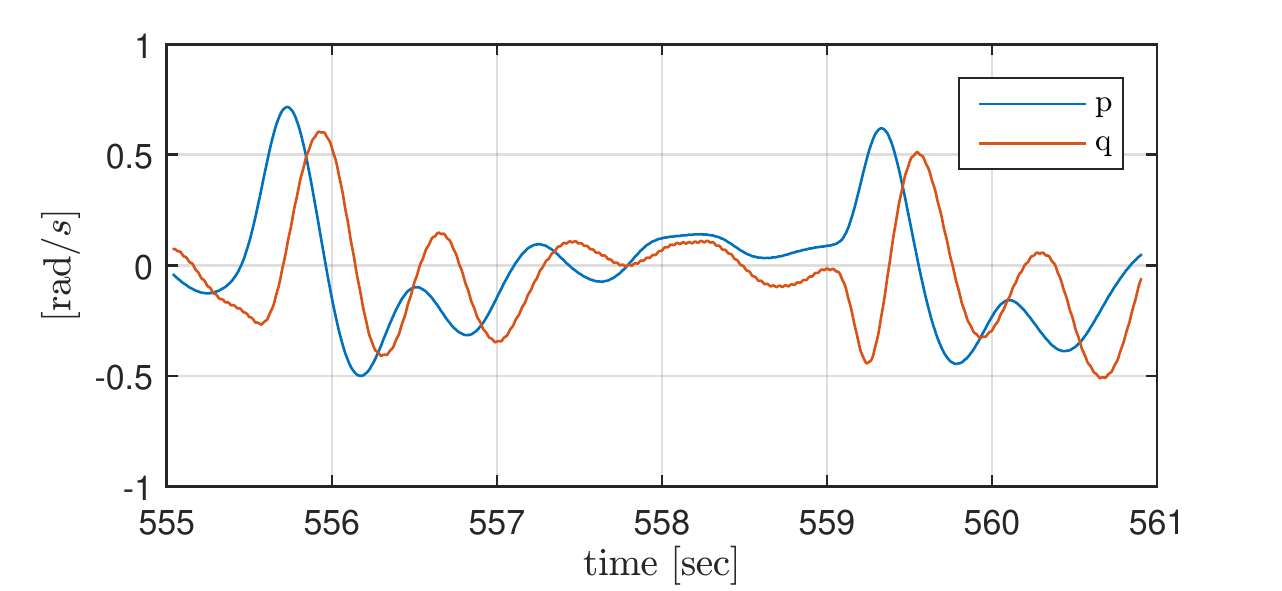}
\caption{An early test flight of \delftacopter with a manually tuned standard helicopter rate controller showed that a doublet step input right ($t=555$ s) and then left ($t=559$ s) on roll rate $p$ yields an undesired but very significant pitch rate $q$. Pilots described this undesired and delayed effect of pitch on roll commands as `\emph{wobbling}'.}
\label{figure:couple_flight}
\end{figure}

Early test flight attempts showed very significant unexplained behavior even in windless indoor hovering flight. Figure~\ref{figure:couple_flight} shows how pitch commands were highly coupled with roll commands and vice verse. To investigate this behavior a theoretic model was derived.

\subsection{Rotor}

To investigate the dynamics of the \delftacopter rotor and fuselage, a simplified model was derived \cite{wagtersmeur2016}. Figure~\ref{figure:RotorModel} illustrates the basic rotor model with rotor radius $R$ and spinning rate $\omega$.
The flapping angle $\beta$ is measured around the spring hinge $K$ and the feathering angle $\theta$ is periodic and follows the setting of the swash-plate cyclic and collective control. The resulting equation of motion of a rotor-blade can be written as:

\begin{figure}[hbt]
\centering
\def\svgwidth{6cm}

\begingroup%
  \makeatletter%
  \providecommand\color[2][]{%
    \errmessage{(Inkscape) Color is used for the text in Inkscape, but the package 'color.sty' is not loaded}%
    \renewcommand\color[2][]{}%
  }%
  \setlength{\unitlength}{\svgwidth}%
  \global\let\svgwidth\undefined%
  \global\let\svgscale\undefined%
  \makeatother%
  \begin{picture}(1,0.86997504)%
\iffinal
    \put(0,0){\includegraphics[width=\unitlength,page=1]{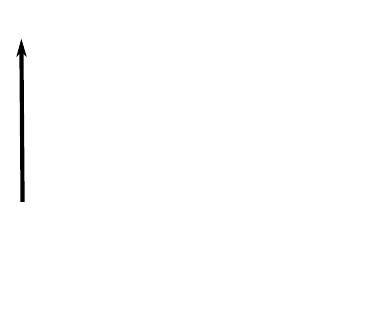}}%
    \put(0,0){\includegraphics[width=\unitlength,page=2]{drawings/blade.pdf}}%
    \put(0,0){\includegraphics[width=\unitlength,page=3]{drawings/blade.pdf}}%
    \put(0,0){\includegraphics[width=\unitlength,page=4]{drawings/blade.pdf}}%
\else
    \put(0,0){\includegraphics[width=\unitlength]{drawings/blade}}%
\fi
    \put(0.01056011,0.81847786){\color[rgb]{0,0,0}\makebox(0,0)[lb]{\smash{$\omega$}}}%
    \put(0.50292194,0.51714643){\color[rgb]{0,0,0}\makebox(0,0)[lb]{\smash{$m$}}}%
    \put(0.34093313,0.30204923){\color[rgb]{0,0,0}\makebox(0,0)[lb]{\smash{$r$}}}%
    \put(0.1262395,0.57115113){\color[rgb]{0,0,0}\makebox(0,0)[lb]{\smash{$K$}}}%
    \put(0.34539838,0.66998191){\color[rgb]{0,0,0}\makebox(0,0)[lb]{\smash{}}}%
    \put(0.56687884,0.10154089){\color[rgb]{0,0,0}\makebox(0,0)[lb]{\smash{$R$}}}%
    \put(0.77453649,0.11587289){\color[rgb]{0,0,0}\makebox(0,0)[lb]{\smash{$\beta$}}}%
  \end{picture}%
\endgroup%

\caption{Simplified rigid rotor model of \delftacopter turning with angular rate $\omega$, mass $m$, radius $R$, spring stiffness $K$ yielding a flapping angle $\beta$.}
\label{figure:RotorModel}
\end{figure}

\begin{equation} \label{equation:flapping}
\ddot{\beta} + \frac{\gamma}{8}\omega\dot{\beta}       + (\omega^{2} + \frac{K}{I}) \beta = \frac{\gamma}{8}\omega^{2}(\theta)
\end{equation}

in which $\gamma$ is the so-called \emph{Lock Number} \cite{bramwell2001bramwell}. The terms in Equation~\ref{equation:flapping} from left to right relate first the inertia of the rotor, its aerodynamic damping, the centrifugal and spring forces to the external excitation.
This concise notation clearly shows that besides rotor rpm $\omega$, the rotor dynamics depend mostly on a single entity \emph{Lock Number} $\gamma$:

\begin{equation}\label{equation:lock}
\gamma = \frac{ \rho  c_{l_\alpha} c R^4 }{ I }
\end{equation}

As given in Equation~\ref{equation:lock}, the \emph{Lock Number} physically contains aerodynamic damping terms (air density $\rho$, chord $c$, rotor radius $R$ and lift coefficient $c_{l_\alpha}$)  divided by flapping inertia $I$. In the \delftacopter design the \emph{Lock Number} is relatively high as the lift coefficient $c_{l_\alpha}$ and radius $R$ are large while the mass and resulting inertia $I$ are very small. 

While Equation~\ref{equation:flapping} shows the importance of rotor inertia in the response rate, it is not sufficient by itself to explain the couplings seen in Figure~\ref{figure:couple_flight}.

\subsection{Fuselage}

\begin{figure}[hbt]
\centering
\def\svgwidth{7cm}
\begingroup%
  \makeatletter%
  \providecommand\color[2][]{%
    \errmessage{(Inkscape) Color is used for the text in Inkscape, but the package 'color.sty' is not loaded}%
    \renewcommand\color[2][]{}%
  }%
  \providecommand\transparent[1]{%
    \errmessage{(Inkscape) Transparency is used (non-zero) for the text in Inkscape, but the package 'transparent.sty' is not loaded}%
    \renewcommand\transparent[1]{}%
  }%
  \providecommand\rotatebox[2]{#2}%
  \ifx\svgwidth\undefined%
    \setlength{\unitlength}{142.49770508bp}%
    \ifx\svgscale\undefined%
      \relax%
    \else%
      \setlength{\unitlength}{\unitlength * \real{\svgscale}}%
    \fi%
  \else%
    \setlength{\unitlength}{\svgwidth}%
  \fi%
  \global\let\svgwidth\undefined%
  \global\let\svgscale\undefined%
  \makeatother%
  \begin{picture}(1,0.54862845)%
	\iffinal
    \put(0,0){\includegraphics[width=\unitlength,page=1]{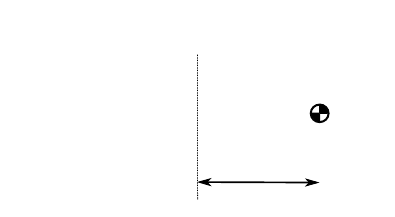}}%
    \put(0,0){\includegraphics[width=\unitlength,page=2]{drawings/bodymodel.pdf}}%
    \put(0,0){\includegraphics[width=\unitlength,page=3]{drawings/bodymodel.pdf}}%
	\else
    \put(0,0){\includegraphics[width=\unitlength]{drawings/bodymodel}}%
	\fi
    \put(0.41194076,0.50485975){\color[rgb]{0,0,0}\makebox(0,0)[lb]{\smash{rotor}}}%
    \put(0.9947806,0.34347338){\color[rgb]{0,0,0}\makebox(0,0)[rb]{\smash{$m_{wing}/4$}}}%
    \put(0.51446784,0.01944485){\color[rgb]{0,0,0}\makebox(0,0)[lb]{\smash{$(l_x,l_y)_{c.g._{wing}}$}}}%
    \put(1.76398035,0.39786431){\color[rgb]{0,0,0}\makebox(0,0)[lb]{\smash{}}}%
    \put(-0.00112598,0.31478115){\color[rgb]{0,0,0}\makebox(0,0)[lb]{\smash{wing}}}%
    \put(0.11775679,0.10304866){\color[rgb]{0,0,0}\makebox(0,0)[lb]{\smash{elevon}}}%
  \end{picture}%
\endgroup%

\caption{Body Model.}
\label{figure:BodyModel}
\end{figure}

To simulate and understand observed pitch and roll couplings a fuselage model is added. Fuselage inertia is playing a crucial role in the control when the fuselage inertia becomes significant compared to the rotor inertia \cite{wagtersmeur2016}. In \delftacopter the weight is spread over the very long wing with a lot of electronics like radios and antennas being placed in the wing tips for electronics reasons. The total weight of \delftacopter is over $4$ kg while a rotor blade is only about $60$ g and the rotor rpm is kept as low as practicable for power reasons. The fuselage inertia can be modeled as shown in Figure~\ref{figure:BodyModel}.

The rotor and body interact with each other in the following ways. Even in case of a fully hinged rotor with $K=0$, when the fuselage rotates, the rotor will automatically follow through the functioning of the swash-plate. In case of non zero $K$ an additional moment will be applied from the fuselage on the rotor whenever they are not in-line. The other way around a moment is transferred from rotor to fuselage through spring $K$ and another moment exists whenever the total lift is not going through the fuselage center of gravity. Forces through the swash-plate linkages are neglected.

\subsection{Simulation}

The combined effect of a light high-lift rotor and heavy non-symmetric fuselage can be clearly visualized in simulation. A model with parameters found in \cite{wagtersmeur2016} is given a step input in pitch $\delta_q$ using a standard helicopter swash-plate as modeled in Equation~\ref{equation:cyclic}.

\begin{equation}\label{equation:cyclic}
\theta = \delta_p \sin(\omega t) + \delta_q \cos(\omega t)
\end{equation}

\begin{figure}[hbt]
\centering
\makebox[\textwidth][c]{\includematlab{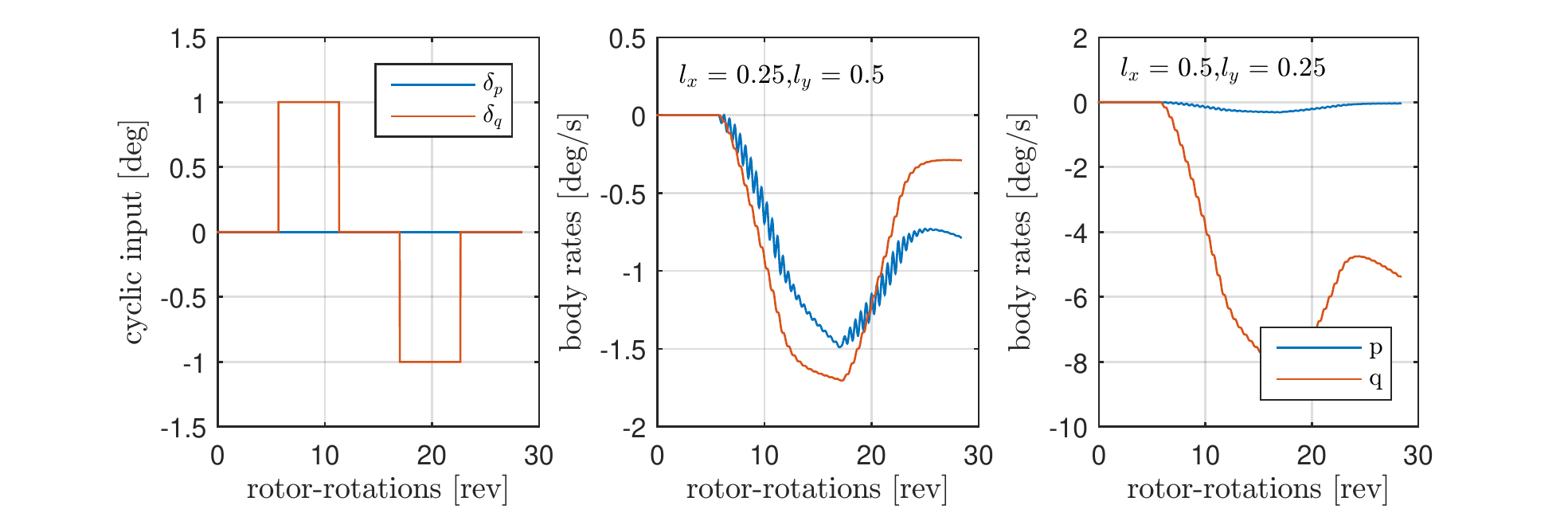}}
\caption{The influence of fuselage inertia on a free body.}
\label{figure:FuselageInertia}
\end{figure}

The resulting cross couplings between pitch and roll for different body inertia are shown in Figure~\ref{figure:FuselageInertia}. A pitch cyclic doublet $\delta_q$ is applied. The simulation results clearly show the resulting desired pitch rate $q$ but show a highly different undesired coupling in roll rate $p$ which it totally different based on the inertia of the fuselage.

In other words, when the rotational inertia of the fuselage is large in pitch and less in roll, then a pitch command on the rotor will start pitching up the rotor plane. The fuselage inertia counteracts this rotation and will result in a pitch down moment on the rotor. The precession of the rotor will turn this into a rolling motion.

A controller for \delftacopter will therefore need to compensate for cross couplings in pitch and roll body motions.

\section{Control}\label{section:Control}

The innerloop control of \delftacopter is an angular rate controller.
The outputs of the rate controller are cyclic commands $\delta_p$ and $\delta_q$, which are mapped to the three servos $\delta_{s1}$, $\delta_{s2}$ and $\delta_{s3}$ that control the swash-plate shown in Figure~\ref{figure:HeadCloseUp} using Equations~\ref{equation:servo1}--\ref{equation:servo3}. 

\begin{eqnarray}
\delta_{s1} &=& \frac{\sqrt{2}}{2} \delta_p - \frac{ \delta_q}{2}\label{equation:servo1}\\
\delta_{s2} &=& - \frac{\sqrt{2}}{2} \delta_p - \frac{ \delta_q}{2}\label{equation:servo2}\\
\delta_{s3} &=& \delta_q\label{equation:servo3}
\end{eqnarray}

Using the on-board SD logging, data was collected in flight to identify the coupled vehicle dynamics. From Section~\ref{section:Rotormodel} the angular acceleration in pitch $\dot{q}$ and roll $\dot{p}$ are expected to result from the cyclic inputs $\delta_p$ and $\delta_q$, the rates in roll $p$ and pitch $q$. An offset $O=1$ is added to the fit to compensate for trim errors.
The control model is shown in Equations \ref{eq:fitp} and \ref{eq:fitq}, where $f_p$ and $f_q$ are linear functions of the parameters.

\begin{equation}
\dot{p} = f_p(O,\delta_x, \delta_y,p,q)
\label{eq:fitp}
\end{equation}

\begin{equation}
\dot{q} = f_q(O,\delta_x, \delta_y,p,q)
\label{eq:fitq}
\end{equation}

\begin{figure}[hbt]
\centering
\includematlab{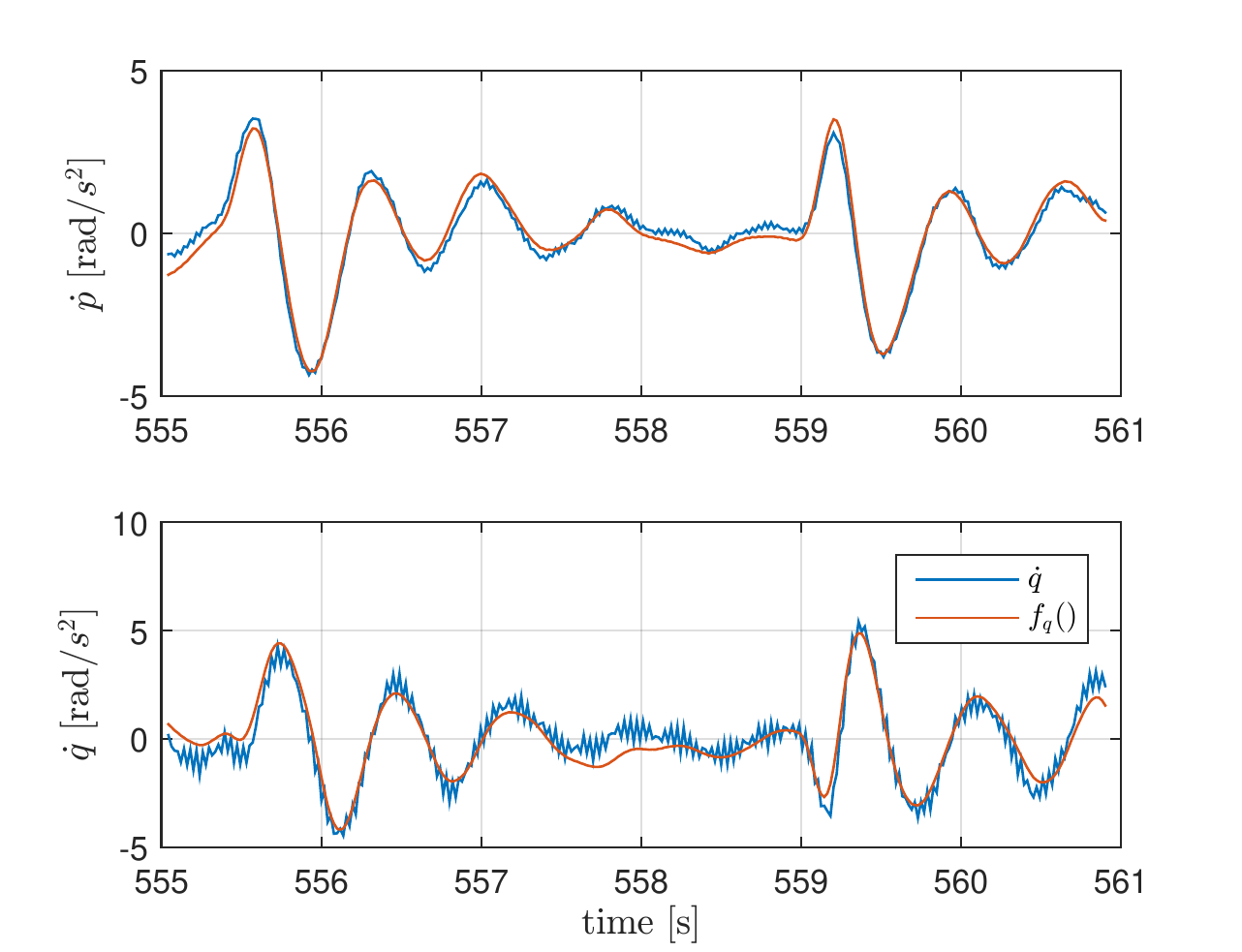}
\caption{Fitting control inputs and body rates to body accelerations. The best model fit relates well to the and the filtered angular acceleration.}
\label{figure:RateFit}
\end{figure}

Figure~\ref{figure:RateFit} shows the angular acceleration in roll and pitch along with the best fit of $f_p()$ and $f_q()$ for a short indoor flight fragment where \delftacopter keeps a constant rpm of $1650$ rpm during a $\approx 20$ degree step in roll.
All signals were filtered with a second order filter with a cutoff frequency of $15$ rad/s.
From the figure it can be seen that this model fit can explain most of the behavior for this part of the flight.
The coefficients that were found using the data shown in Figure~\ref{figure:RateFit} are given in Table~\ref{table:id_param}.

\begin{table}[hbt]
\caption{Identified parameters.}
\label{table:id_param}
\begin{center}
\begin{tabular}{|l|c|c|}
\hline
Coefficient & $f_p$ & $f_q$\\ \hline
\hline
$C_O$ & -2.4661 & -2.8847 \\ \hline
$C_{\delta_x}$ & 0.0032 & -0.0044\\ \hline
$C_{\delta_y}$ & 0.0011 & 0.0073\\ \hline
$C_{p}$ & -0.5703 &  7.4479\\ \hline
$C_{q}$ & -3.4308 & -3.4487\\ \hline
\hline
\end{tabular}
\end{center}
\end{table}

When looking closely at the coefficients for $C_p$ and $C_q$ in Table~\ref{table:id_param}, they confirm that a roll rate causes a pitch acceleration and vice verse.

Taking into account the identified couplings, the linear controller is revised to:

\begin{equation}
\left[ \begin{array}{c}
\delta_x \\
\delta_y 
\end{array} \right]
= G^{-1} 
\left[ \begin{array}{c}
K_p \cdot p_{err} + q \cdot C_{q_{\dot{p}}} \cdot K_c \\
K_q \cdot q_{err} + p \cdot C_{p_{\dot{q}}} \cdot K_c
\end{array} \right]
\label{eq:final_control}
\end{equation}

Where $p_{err}$ and $q_{err}$ are the difference between the pilot command and the actual rates of the vehicle, and

\begin{equation}
G = 
\left[ \begin{array}{cc}
C_{{\delta_{x_{\dot{p}}}}} & C_{{\delta_{y_{\dot{p}}}}} \\
C_{{\delta_{x_{\dot{q}}}}} & C_{{\delta_{y_{\dot{q}}}}}
\end{array} \right]
\label{eq:G}
\end{equation}

An in-flight tuning parameter $K_c$ is introduced with a value between 0 and 1. 
It was introduced in order to gradually enable the compensation of angular acceleration due to rates.
Test flights showed that a value of $K_c=0.5$ gives better results than a value of $K_c=1$.
This may be caused by actuator dynamics, as a control moment can not be instantly generated when a rate is measured.
More research is necessary to better explain why $K_c=1$ still gives a wobble.

\begin{figure}[hbt]
\centering
\includematlab{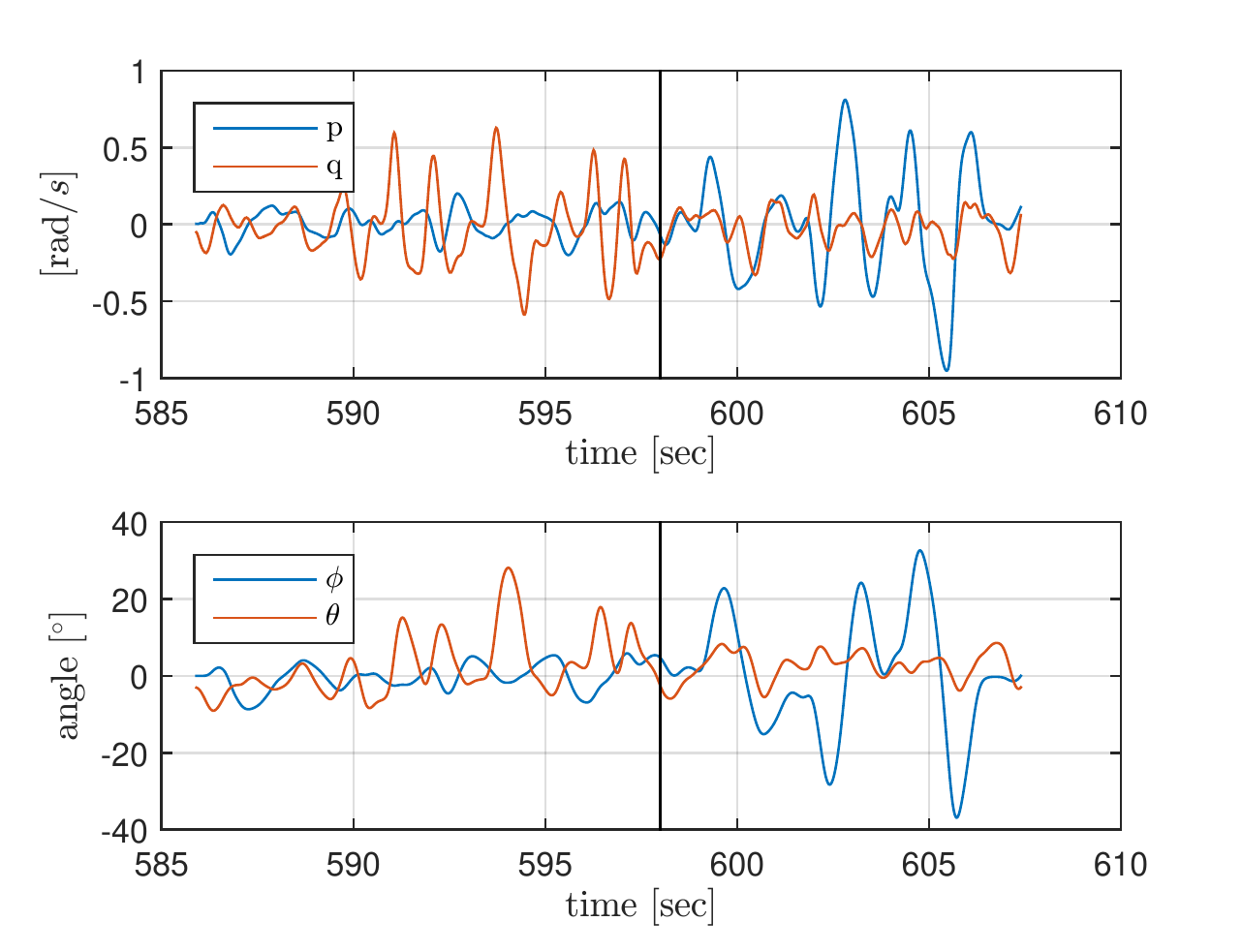}
\caption{Flight with $K_c=0.5$.}
\label{fig:final_flight}
\end{figure}

Figure~\ref{fig:final_flight} shows the measured angular rates of the vehicle during some pitch maneuvers in the first part of the flight and some roll maneuvers in the second part of the flight. 
The rates were filtered with a second order filter with a cutoff frequency of $25$ rad/s.
From Figure~\ref{fig:final_flight} it can be seen that no wobble is present, and the motion in roll and pitch is uncoupled.
When compared back to the initial situation in Figure \ref{figure:couple_flight}, one can see that the control was highly improved.

\section{Vision}\label{section:Vision}

\begin{figure}[htb]
\centering
\includegraphics[width=8cm]{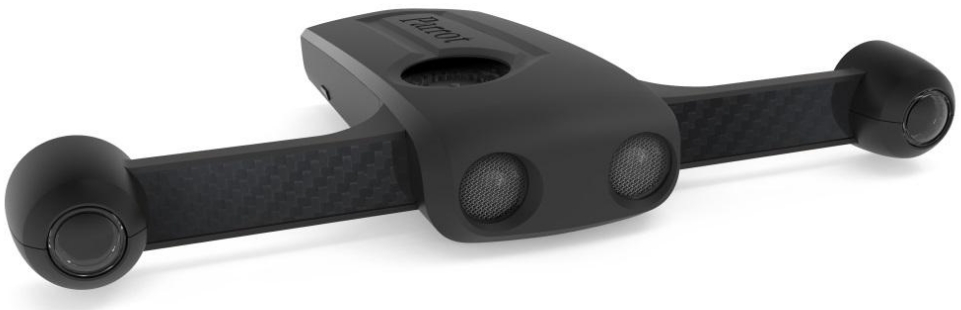}
\caption{The S.L.A.M.dunk vision system.}
\label{figure:slamdunk}
\end{figure}

\delftacopter was equipped with a state of the art computer vision system as can be seen in Figure~\ref{figure:slamdunk}: a prototype of the Parrot \slamdunk.

\begin{figure}[h]
\centering
\includegraphics[width=10.4cm]{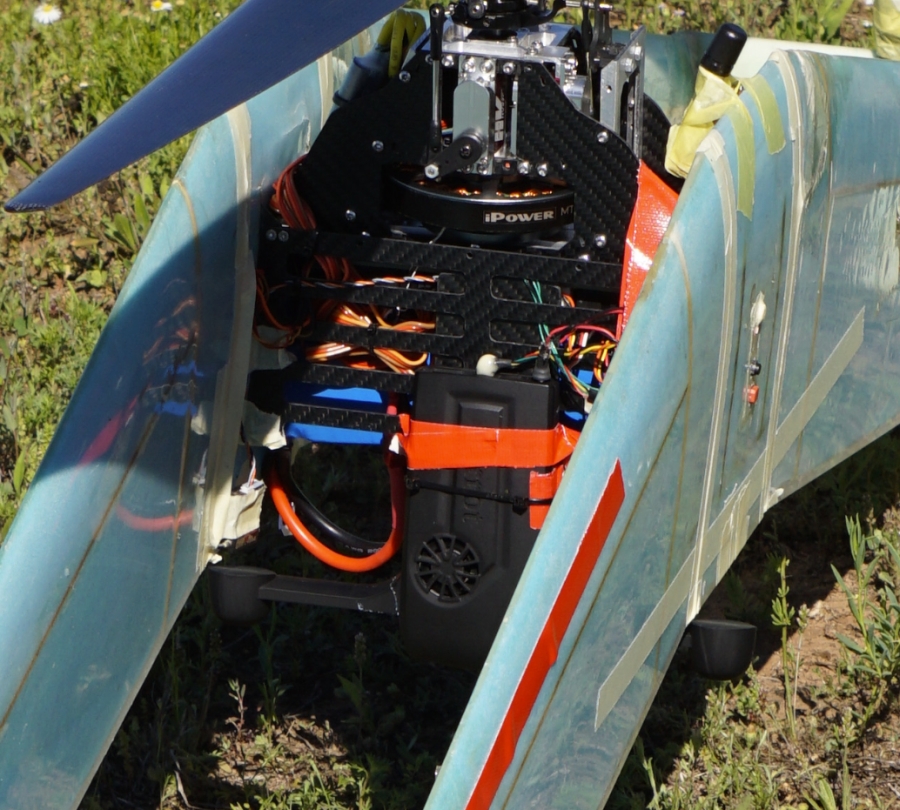}
\caption{The \slamdunk vision system was mounted on \delftacopter looking away from the main rotor. In hovering flight this means the camera have a perfect view of the ground, and in forward flight the camera looks backwards. Thanks to the wide-field-of-view fish-eye lenses the camera can nevertheless still look vertically down during forward flight.}
\label{figure:slamdunk_position}
\end{figure}

Our prototype\footnote{The unit used in \delftacopter was a prototype product in development, the \slamdunk will have improved specifications at release time.} \slamdunk API delivered a 96x96 pixels depth map which, if overlaid over the original camera image, was the center 640x480 pixel range, cropped from the full 1280x1024 camera resolution. The depth map was generated at $30$ fps, by means of Semi Global Matching accelerated by the GPU. On \delftacopter the \slamdunk looked straight down towards the ground when in hover mode, as can be seen in Figure~\ref{figure:slamdunk_position}.

\begin{table}[ht]
\caption{\slamdunk prototype properties.}
\label{table:slamdunk}
\begin{center}
\begin{tabular}{|l|l|}
\hline
  Processor & Nvidia Tegra TK1 \\ \hline
  Cameras & 1280x1024 RGB\\ \hline
  Depth map & 640x480 \\ \hline
  Frame rate & 30 fps \\ \hline
  Stereo base line  & $20$ cm\\ \hline
  Lenses  & Fish eye\\ \hline
  Sensors & 10-DOF IMU and sonar \\ \hline
\end{tabular}
\end{center}
\end{table}

This system was used for: 1) detection of Joe, 2) landing spot selection, 3) obstacle avoidance during landing and 4) determining the moment of touchdown. Lastly, a feature was devised to automatically calibrate the attitude error between the airframe and the \slamdunk.
The hardware specifications are shown in Table~\ref{table:slamdunk}.

\subsection{Find Joe}

The competition organizers only provide an approximate GPS position of bushwalker Joe ($\pm 100$ meters accuracy) \cite{outbackrules} while strict requirements were imposed on landing locations. Therefore the vision system needed to search and pinpoint the exact position of Joe.
The Medical Outback challenge rules dictate to keep a minimum distance of $30$ meters from Joe in all directions, in order to comply with CASA (airspace regulatory body in Australia) regulations.
In practice this meant \delftacopter needed to maintain a minimum height of $40$ meters while searching in order to account for height measurement errors and gusts.
Joe was to be visible in a field near a farm, wearing a normal blue jean and an Australian Akubra head.

Due to limitations imposed by the fish-eye lens, the overall quality of the images of Joe taken from a moving and vibrating platform at $40$ meters was low due to blur and low spatial resolution.
Joe was reported to be standing upright, which is why a birds-eye view could be advantageous to detect pose and human parts features \cite{rudol2008human}.
This non perpendicular view resulted in an increased observation distance to Joe, and even smaller pixel representations. Moreover, no example data of the Australian scene or Joe was available beforehand to us, complicating methods that need many training examples \cite{de2016towards}, and due to ongoing development on \delftacopter itself, vision test flights on the final platform were scarce.
Therefore, emphasis was placed on an algorithm simple enough to test and tune at the last moment on the Australian scene. Instead of using a real human for Joe, a full-size dressed dummy was used. This means no movement or thermal features could be used \cite{gaszczak2011real}.
Instead, a simple color plus shape filter was used as a salience detector to detect possible Joes, which were clustered based on there projected GPS locations.
The best exemplary thumbnails of each cluster were selected, cut out, and updated as a better view angle become available.
These thumbnails, accompanied with their projected GPS positions and Joe likelihood scores (how many consecutive frames Joe could be detected at a given location) were calculated, and sent over the data link to the ground control station.
A video of the Joe detection algorithm in action can be viewed online \footnote{Joe detection: \url{https://www.youtube.com/watch?v=GVTHuwg3VJY}}.

\begin{figure}[ht]
\centering
\includegraphics[width=9cm]{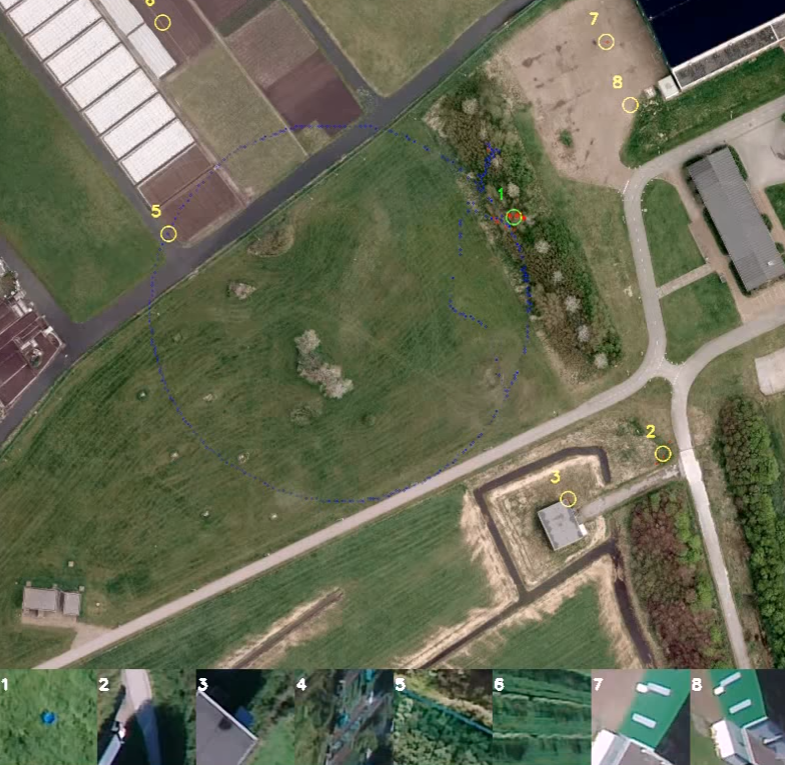}
\caption{Person Detection System.}
\label{figure:vision:joe}
\end{figure}

The vision operator was shown an interface with a map, numbers and list with accompanying thumbnails as shown in Figure~\ref{figure:vision:joe}.
The blue dots show the flight path, the red dots show Joe sightings, the yellow circles shows the average location of the clustered red dots and the green circle shows the winning Joe sighting based on the highest Joe likeliness score.
During the competition flight, a flight pattern in the shape of a \emph{W} was pre-programmed to cover the whole possible area.


\subsection{Selecting a landing spot}

\begin{figure}[ht]
\centering
\includegraphics[width=6cm]{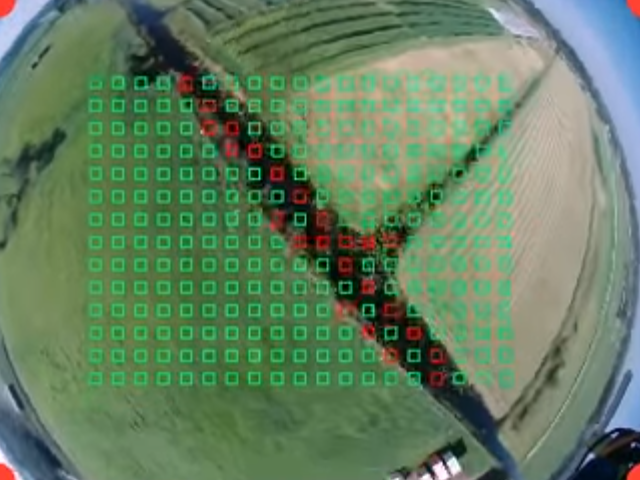}
\caption{Landing site classification into land-able and dangerous based on texture comparison.}
\label{figure:vision:patches}
\end{figure}

During the search for Joe, the surface was analyzed at the same time for landibility.
Textures of areas with a known good surface (farmers land, grass, desert, road, etc) and bad surface (water, trees, roofs, etc) were annotated beforehand, and during flight classified with a simple Euclidean distance texture comparison on sub-sampled patches from the video stream as shown in Figure~\ref{figure:vision:patches}.
A video of the on-board results can be viewed online\footnote{Landing spot classification: \url{https://www.youtube.com/watch?v=RQ6F2ccMv8g}}.
On this result, a blob-finder was applied and the biggest blob close to the selected Joe was automatically converted to the landing waypoint.

\subsection{Obstacle avoidance}

The landing site is in a natural, unstructured and unfamiliar environment which means the likelihood of unforeseen obstacles such as houses or trees is high. Some assumptions on the environment are made. It is assumed that in the targeted area the surface is commonly flat enough for the \delftacopter to land. Furthermore, although obstacles are assumed to be of frequent occurrence, they are assumed to be sparsely located such that enough room exists between (small groups of) obstacles to land. 
During hover, the \slamdunk is looking downwards and an algorithm based on the depth map provided by the \slamdunk is implemented to land safely. The depth-map can detect obstacles up to $\approx30$ m away but has significant noise in the farthest $10$ m. 
To prevent false positives from the depth map the landing is only enabled at heights lower then $20$ m above ground level (AGL).
The IMU is information is used to determine the pixel location in the image that is straight down.
This is especially important in the \delftacopter design that can fly at considerable bank angles in hover in windy conditions.
During landing a circular area in the image around the current straight down pixel is selected.
A moving average depth is calculated in this area in order to determine the closest distance, the average distance and the minimum distance.
Flatness is defined as the closeness between the minimum and maximum to the average.
When the flatness is sufficiently high according to a threshold, the surface is considered safe for further fast descend.
Otherwise, the aircraft is re-positioned towards the area with the greatest depth. 
%
%
A proportional gain is steering \delftacopter laterally away from the global minimum of the moving average depth around the aircraft while the descend speed is decreased. 
Several tests showed \delftacopter to be able to avoid all obstacles visible in stereo vision depth map during landing, like for instance trees and structures.
A video of the on-board results can be viewed online \footnote{Landing obstacle avoidance: \url{https://www.youtube.com/watch?v=2eCi8VJiDcs}}.


\begin{figure}[ht]
\centering
\includegraphics[width=6cm]{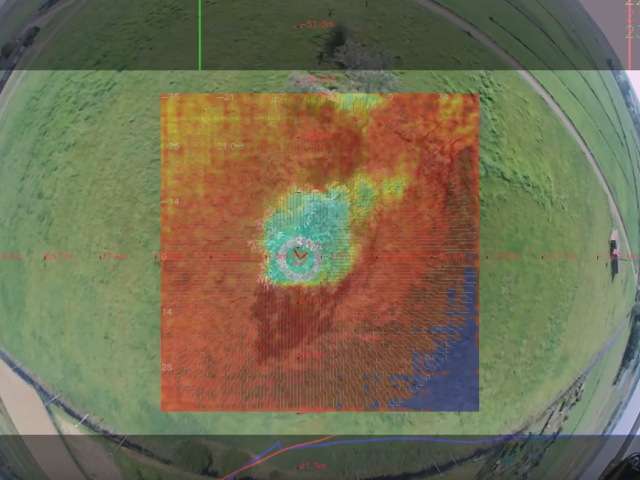}
\caption{Landing avoidance system based on \slamdunk disparity map shown in color shading in the center clearly showing the high tree in the image.}
\label{figure:vision:landing}
\end{figure}

\subsection{Touch down}

To make repeatable precise and smooth landings, predicting the exact moment of touchdown is very important.
The \slamdunk can measure the height to the ground from over $20$ m down to $10$ cm by mixing the sonar and the stereo depth.
This precise height combined with \delftacopter attitude information from the autopilot provides the required information to time landings successfully repeatedly.
A video of a full autonomous landing can be viewed online\footnote{Full autonomous flight: \url{https://www.youtube.com/watch?v=8aakE1WlQQ0}}.

\subsection{Attitude error calibration}
The \slamdunk was designed to be removable in order to allow easy access to the battery and electronics inside \delftacopter.
This can cause a possible discrepancy in the attitude between the camera and the airframe, depending on the mounting process.
However, an exact attitude measurement is necessary in order to precisely geo-locate obstacles and Joe.
Using the difference between the IMU embedded in the \slamdunk and the IMU of the autopilot, the offset in attitude is determined automatically during the start-up phase.
This lowers the requirements on the mounting system significantly.

\section{Flight}\label{section:Flight}

\subsection{Transitioning flight}

\begin{figure}[h!!!!!!!!!!!!!!!!!!!!!!!!!!!!!!!!!!!!!!!!!!!!!!!!!!!!!!!!!!!!!!!!!!!!!!!!]
\centering
\makebox[\textwidth][c]{\includematlab{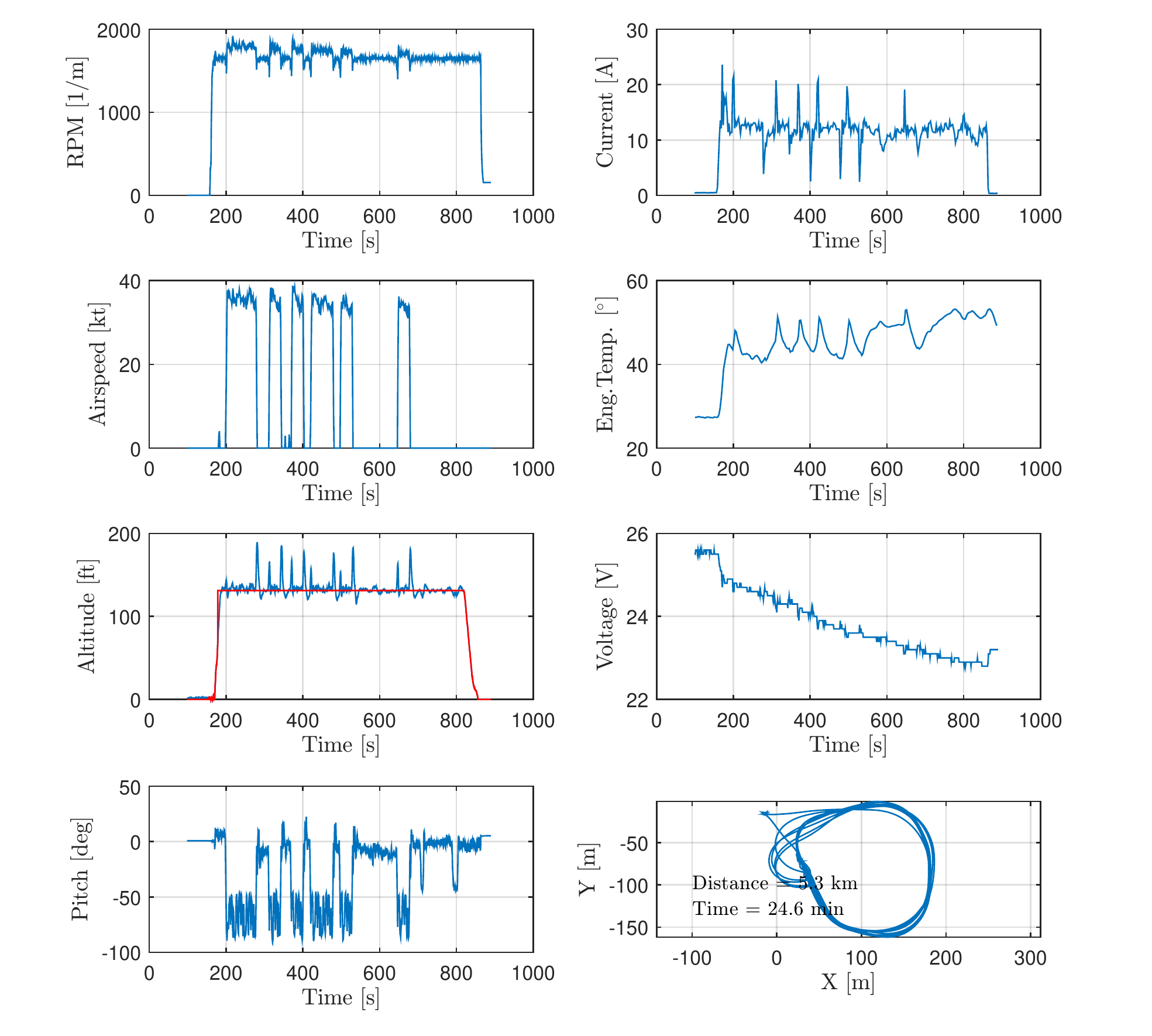}}
\caption{Test flight containing six transitions from hover to forward flight and back.}
\label{figure:Flight:ManyTransitions}
\end{figure}

Figure~\ref{figure:Flight:ManyTransitions} shows a flight with 6 transitions from hover to forward flight and back. 
During a transition from hover to forward, \delftacopter has a small drop in altitude of about $10$ ft.
During the transition back from fast forward flight back to hover a much more significant altitude overshoot is observed of about $60$ ft due to excess energy during the fast transition.
The flight is performed in a very confined area of about $150$ by $150$ and in forward flight \delftacopter is turning most of the time.
Figure~\ref{figure:Flight:ManyTransitions} also shows how during every hover the engine temperature is rising due to the increased load.

\subsection{Efficiency Testing}

\begin{figure}[hbtp]
\centering
\makebox[\textwidth][c]{\includematlab{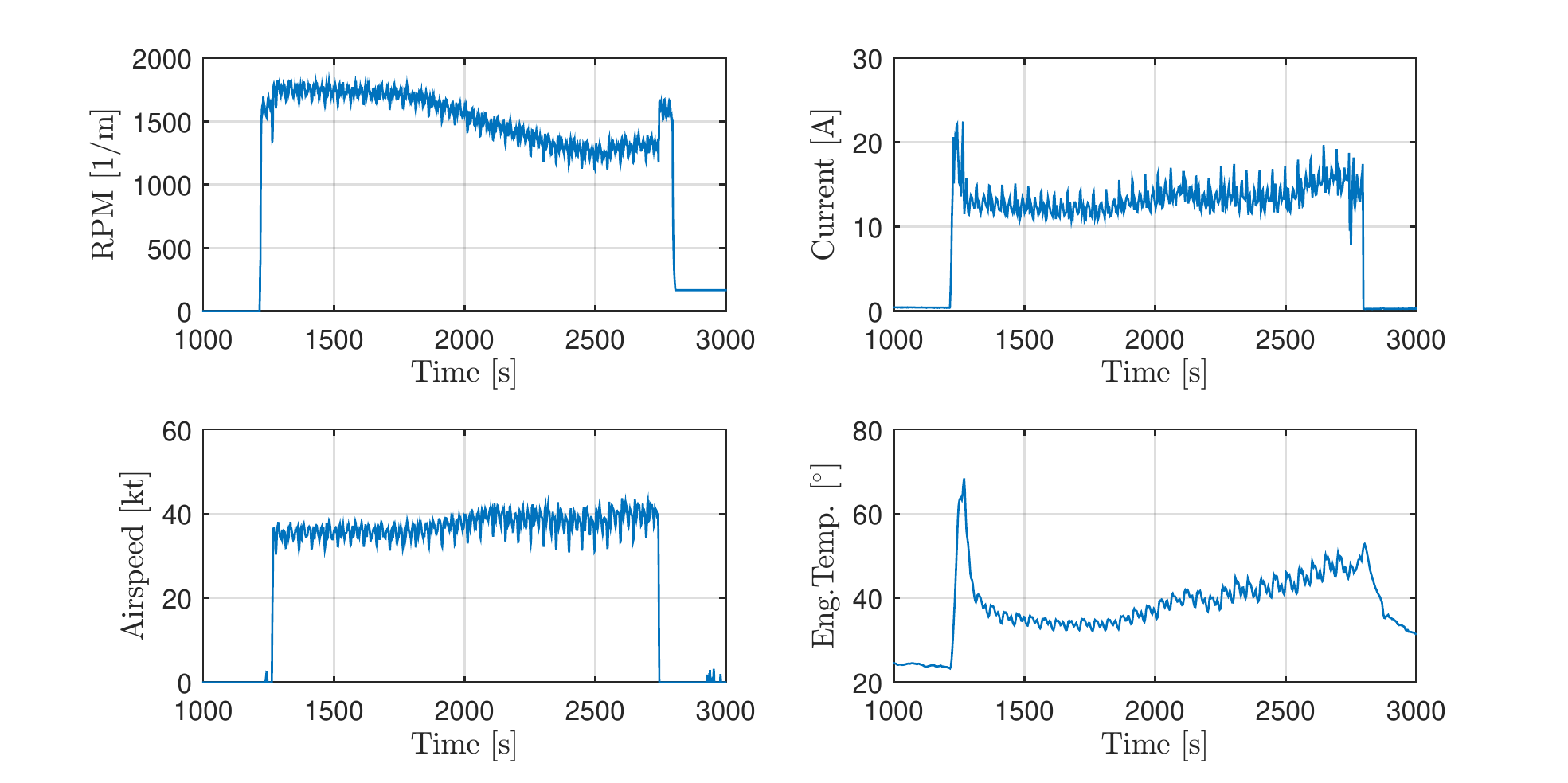}}
\caption{In search for the optimal rotor RPM for most efficient forward flight, a long outdoor flight on a very calm evening was performed in which a large range of pitch and throttle settings were tested.}
\label{figure:Flight:ReduceHeadSpeed}
\end{figure}

During another test flight shown in Figure~\ref{figure:Flight:ReduceHeadSpeed} an attempt is made to find the most optimal forward flight regime.
The design from Section~\ref{section:Propulsion} was suggesting that lower rotor RPM in forward flight should be more efficient.
The actual flight data does however not clearly show an efficiency increase.
This corresponds exactly to the windtunnel observations from Section~\ref{section:Windtunnel}.
The rising motor temperature shows that the high motor load does decrease the electrical efficiency.
Since no loss of total efficiency is observed, this means the propeller efficiency indeed increases but is undone by the loss of electric motor efficiency.

\subsection{Competition Flight}

\begin{figure}[hbtp]
\centering
\includematlab{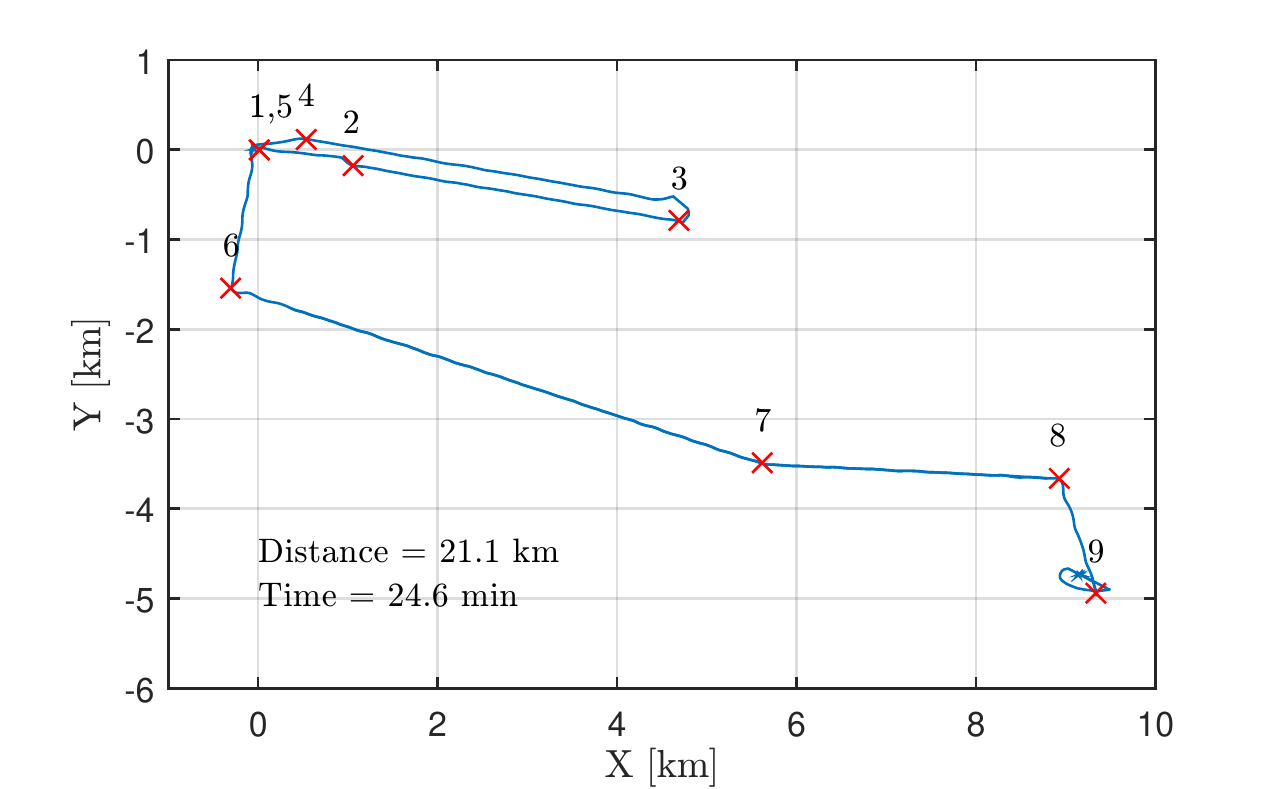}
\caption{Competition flight ground track. The flight starts at waypoint $1$ and searches for Joe around waypoint $9$. Including the hovering take-off and landing the flight to Joe took $24.6$ min and the total distance of the flight was $11.4$ Nm or $21.1$ km.}
\label{figure:Flight:Competition}
\end{figure}

\begin{figure}[hbtp]
\centering
\makebox[\textwidth][c]{\includematlab{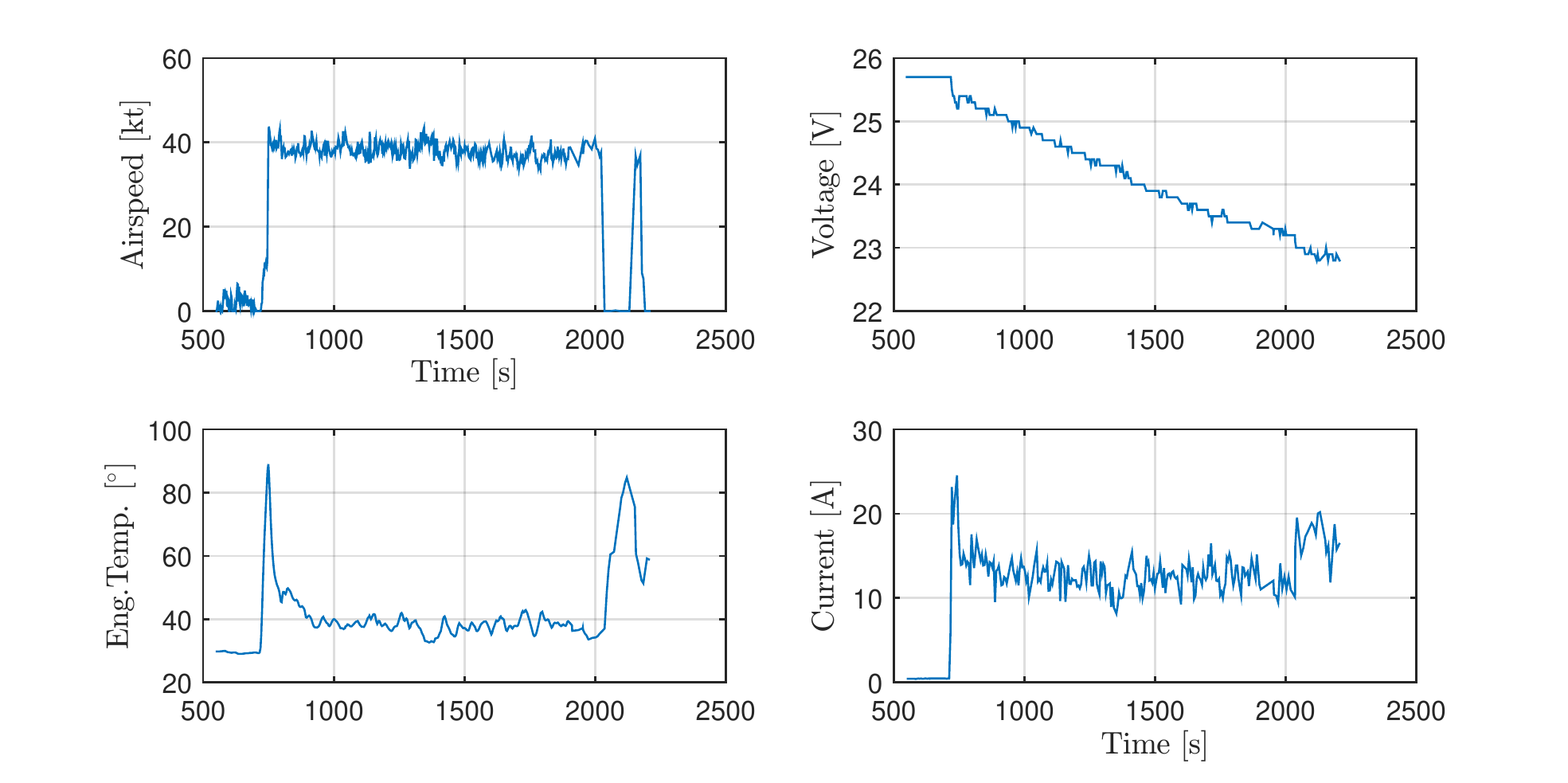}}
\caption{Competition flight data of the flight to Joe. The average airspeed during the flight as roughly $40$ kt or $21.5$ m/s.
During hover the engine temperature rises to $80^\circ$ but during forward flight it settles at a value of about $40^\circ$. The current during the climbing hover in the first phase is about $23$ A or $540$ Watt, during the cruise the current reduces to about $12$ A or $280$ Watt.
Non-climbing hover is achieved at $18$ A or $420$ Watt.}
\label{figure:Flight:CompetitionData}
\end{figure}

A top view of the competition flight from take-off at point 1 to the destination at point 9 is shown in Figure~\ref{figure:Flight:Competition}. The total traveled distance to Joe is $11.4$ Nm or $21.1$ km one-way. Figure~\ref{figure:Flight:CompetitionData} depicts the corresponding flight parameters of the competition flight. \delftacopter was carrying a $10.5$ Ah 6-cell Lithium-Polymer battery pack of $1.6$ kg.

\smallskip

\section{Conclusion} \label{section:Conclusions}

\subsection{Concept}

The \delftacopter concept was selected for its efficient hovering with one large rotor, control authority in hover with fast cyclic control, simple and structurally strong biplane delta-wing design that also serves as landing gear and yields improved stall behavior over single wings. The biplane also reduces the lateral surface affected by turbulence and wind during hover of this tail-sitter VTOL aircraft. \delftacopter was built, tested in a windtunnel setup and in-flight and participated in the \emph{Outback Medical Express Challenge 2016} where it won the seconds prize\footnote{\url{http://www.delftacopter.nl/}}.

The concept is applicable to a variety if scenarios, especially when long-distance and efficient hovering at minimal weight are driving requirements. 


\subsection{Propulsion}

The hover efficiency of \delftacopter is high thanks to it large single low rpm optimized rotor. The energy efficiency during fast forward flight turned out to be lower than expected from computations. This is most likely due to higher than expected airframe drag of the final built combined with significantly reduced motor efficiency at high torque.

\subsection{Control}

The cyclic control of \delftacopter yields very fast and powerful attitude control. But without taking into account the rotor-fuselage dynamics, unacceptable couplings were observed. 
When designing hybrid aircraft mixing conventional cyclic controlled helicopters and fixed wings, it is crucial to understand the interactions between rotor and wing in order to optimize the design. The \emph{Lock Number} and rotor hinge spring stiffness $K$ where shown to influence the speed of the rotor response.
When adding a high inertial fuselage to the model, it showed the same type of behavior in simulation as the real \delftacopter.
The non-homogeneous inertia of the fuselage and fuselage-rotor interactions adds non-symmetrical couplings between the pitch and roll axes and affect the direction of the control effectiveness of the swash-plate.
Compensation for the above effects was derived an converted to a controller strategy to remove unwanted couplings sufficiently for flight.

\subsection{Vision}

The stereo depth map properties of the \slamdunk\footnote{\slamdunk from \url{https://www.parrot.com/}} allow nice planning around obstacles and selection of flat areas during landing. But because of the very wide field-of-view fish-eye lenses, the resolution of the images when flying at cruise altitude is low. This makes reliable fully autonomous search for persons at a higher altitude very difficult. A salient detector with human feedback option was developed to automate the mission as much as possible while allowing human validation of the selected target.

\section{Recommendations}\label{section:Recommendations}


The current design has shown outstanding efficiency in hovering flight and slightly less efficiency than predicted during forward flight.
During the Outback Medical Express Challenge but also most other application scenarios of \delftacopter, the forward flight phase is the predominant mode of flight. 

A future design could therefore place slightly less emphasis on hover efficiency and more on forward flight where some efficiency gains highly affect the operational range and flight speed.
A smaller diameter rotor/propeller will also put less demand on the torque requirements of the direct drive motor which was currently overloaded in fast forward flight.

When higher torque direct drive brushless motors become available, they could also yield efficiency improvements in fast forward flight.

\section*{Acknowledgments}

We thank the sponsors of the TU-Delft Outback Medical Challenge Entry to have made this research possible.

\section*{Appendix}\label{section:Appendix}

\begin{figure}[p]
\centering
\def\svgwidth{0.9\linewidth}
\begingroup%
  \makeatletter%
  \providecommand\color[2][]{%
    \errmessage{(Inkscape) Color is used for the text in Inkscape, but the package 'color.sty' is not loaded}%
    \renewcommand\color[2][]{}%
  }%
  \providecommand\transparent[1]{%
    \errmessage{(Inkscape) Transparency is used (non-zero) for the text in Inkscape, but the package 'transparent.sty' is not loaded}%
    \renewcommand\transparent[1]{}%
  }%
  \providecommand\rotatebox[2]{#2}%
  \ifx\svgwidth\undefined%
    \setlength{\unitlength}{626.96923254bp}%
    \ifx\svgscale\undefined%
      \relax%
    \else%
      \setlength{\unitlength}{\unitlength * \real{\svgscale}}%
    \fi%
  \else%
    \setlength{\unitlength}{\svgwidth}%
  \fi%
  \global\let\svgwidth\undefined%
  \global\let\svgscale\undefined%
  \makeatother%
  \begin{picture}(1,1.47384382)%
	\iffinal
    \put(0,0){\includegraphics[width=\unitlength,page=1]{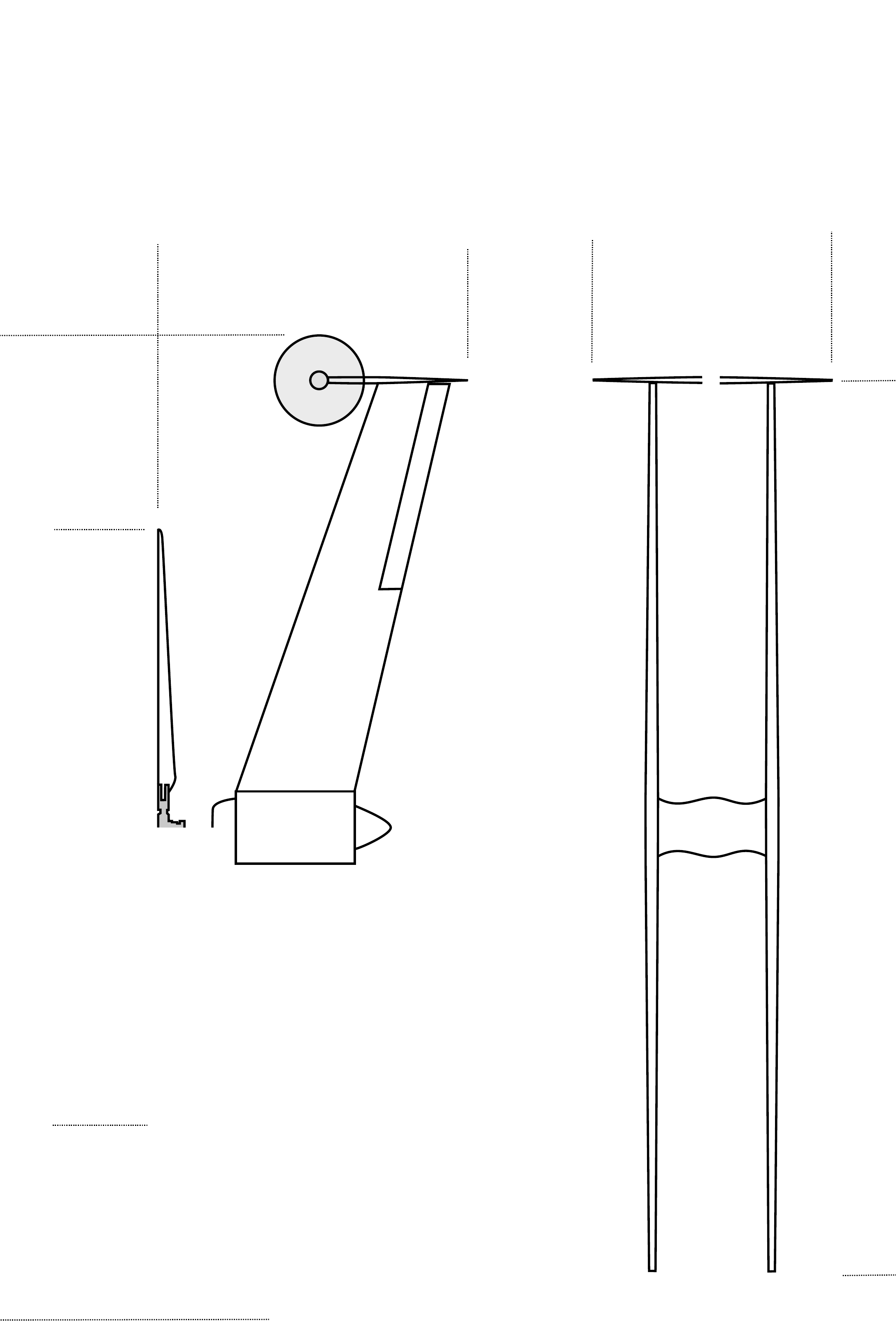}}%
    \put(0,0){\includegraphics[width=\unitlength,page=2]{drawings/topview.pdf}}%
    \put(0,0){\includegraphics[width=\unitlength,page=3]{drawings/topview.pdf}}%
		\put(0,0){\includegraphics[width=\unitlength,page=4]{drawings/topview.pdf}}%
	\else
    \put(0,0){\includegraphics[width=\unitlength]{drawings/topview}}%
	\fi
    \put(0.61509101,0.54832289){\color[rgb]{0,0,0}\rotatebox{90}{\makebox(0,0)[b]{\smash{$10$ cm}}}}%
    \put(0.96170613,0.54365736){\color[rgb]{0,0,0}\rotatebox{90}{\makebox(0,0)[b]{\smash{$150$ cm}}}}%
    \put(0.79720747,1.16561124){\color[rgb]{0,0,0}\makebox(0,0)[b]{\smash{$40$ cm}}}%
    \put(0.34538941,1.16172575){\color[rgb]{0,0,0}\makebox(0,0)[b]{\smash{$52$ cm}}}%
    \put(0.0855674,0.54456995){\color[rgb]{0,0,0}\rotatebox{90}{\makebox(0,0)[b]{\smash{$100$ cm}}}}%
    \put(0.02918293,0.54944272){\color[rgb]{0,0,0}\rotatebox{90}{\makebox(0,0)[b]{\smash{$165$ cm}}}}%
    \put(0.62245705,1.34045106){\color[rgb]{0,0,0}\rotatebox{90}{\makebox(0,0)[b]{\smash{$40$ cm}}}}%
    \put(0.29284734,0.20506917){\color[rgb]{0,0,0}\makebox(0,0)[b]{\smash{$14$ cm}}}%
  \end{picture}%
\endgroup%

\caption{3D View of \delftacopter. The center part consists of a 480-sized helicopter head with direct drive motor. Two pairs of delta-wings are attached to form a box structure. The wing tips are stabilizing the delta-wing by increasing its $C_{m_\beta}$, they are connecting the top and bottom wing and providing structural strength, they are housing communication modems and acting as landing gear.}
\label{figure:Appendix:Topview}
\end{figure}

\begin{figure}[hbt]
\centering
\includegraphics[width=7cm]{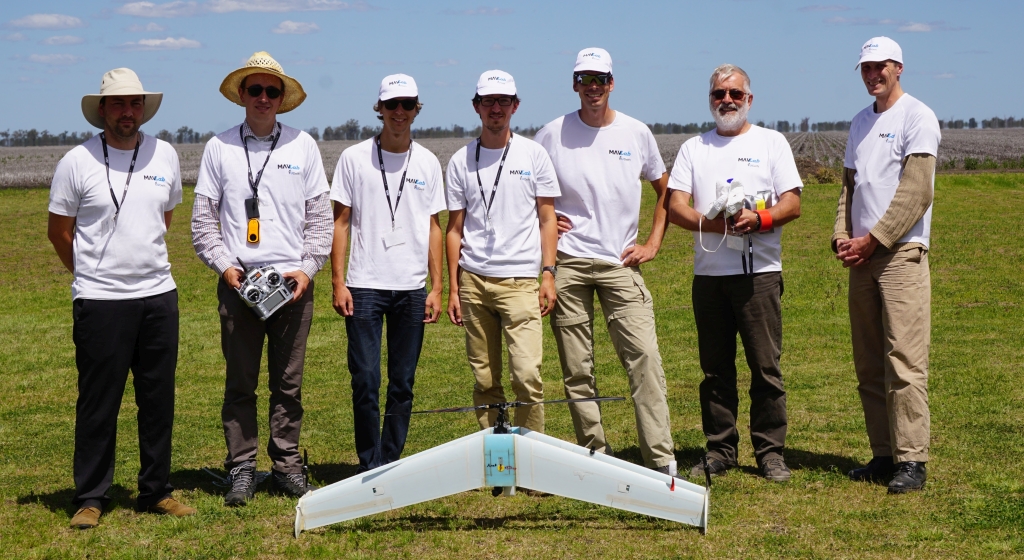}
\caption{\delftacopter team at the \emph{Outback Medical Express UAV Challenge 2016}, Dalby Australia.}
\label{figure:team}
\end{figure}

Figure~\ref{figure:Appendix:Topview} shows the three views of \delftacopter with sizing information. Key specification values of \delftacopter are repeated in Table~\ref{table:specs}. Finally Figure~\ref{figure:team} gives an overview of the team.

\begin{table}[hbt]
	\centering
	\caption{\delftacopter key specifications}
	\label{table:specs}
		\begin{tabular}{|l|l|}
           \hline
			Property & Value \\
			\hline
            \hline
Weight & $4$ kg \\
MTOW & $4.5$ kg \\
Weight & $4$ kg \\
Wing Area & $0.496$ $m^2$ \\
Wing Loading & $8$ $kg/m^2$ \\		  
Span & $1.54$ m \\
Length & $0.6$ m \\
Height & $0.4$ m \\
\hline 
Cruise speed & $40$ kt at $150$ Watt \\
Most efficient speed & $35$ kt at $120$ Watt \\
Maximum speed & $58$ kt \\
\hline
Main battery & $10.5$ Ah 6 Cell LiPo \\
FTD battery & $250$ mAh 2 Cell LiFe \\
\hline
		\end{tabular}
\end{table}

\bibliographystyle{unsrt}
\bibliography{outback_bibliography}

\end{document}